\documentclass[journal]{IEEEtran}
\usepackage{cite}
\usepackage{amsfonts}
\usepackage{color}
\usepackage{booktabs}
\usepackage{multirow}
\usepackage{amssymb}

\ifCLASSINFOpdf
\else
\fi
\usepackage{amsmath}
\usepackage{graphicx}
\usepackage{algorithmic}
\usepackage{titletoc}
\usepackage{threeparttable}

\begin{document}
%
\title{NFANet: A Novel Method for Weakly Supervised Water Extraction from High-Resolution Remote Sensing Imagery}
%
%
%

\author{Ming Lu, Leyuan Fang,~\IEEEmembership{Senior Member, ~IEEE}, Muxing Li, Bob Zhang,~\IEEEmembership{Senior Member, ~IEEE}, Yi Zhang,~\IEEEmembership{Senior Member, ~IEEE}, and Pedram Ghamisi, ~\IEEEmembership{Senior Member, ~IEEE}
	
	\thanks{This work was supported in part by the National Natural Science Fund of China under Grant 61922029, in part by the Science and Technology Plan Project Fund of Hunan Province under Grant 2019RS2016, in part by the Key Research and Development Program of Hunan under Grant 2021SK2039, and in Part by the Natural Science Fund of Hunan Province under Grant 2021JJ30003.( Corresponding author: Leyuan Fang. )}

	\thanks{Ming Lu is with the College of Electrical and Information Engineering, Hunan University, Changsha 410082, China, and also with the State Key Laboratory of Integrated Services Networks, Xidian University, Xian, 710126, China (e-mail: 1148462196@qq.com).}
	
	\thanks{Leyuan Fang is with the College of Electrical and Information Engineering, Hunan University, Changsha 410082, China, and also with the Peng Cheng Laboratory, Shenzhen 518000, China (e-mail: fangleyuan@gmail.com).}
	
	\thanks{Muxing Li is with the College of Engineering \& Computer Science, Australian National University, ACT 2601, Australia (e-mail: u6481081@anu.edu.au) }
	
	\thanks{Bob Zhang is with the Department of Computer and Information Science, University of Macau, Macau 999078, China (e-mail: bobzhang@um.edu.mo)}
	
	\thanks{Y. Zhang is with the College of Computer Science, Sichuan University, Chengdu 610065, China (e-mail: yzhang@scu.edu.cn).}
	
	\thanks{Pedram Ghamisi is with the Helmholtz-Zentrum Dresden-Rossendorf (HZDR), Helmholtz Institute Freiberg for Resource Technology, 09599 Freiberg, Germany, and also with the Institute of Advanced Research in Artificial Intelligence (IARAI), 1030 Vienna, Austria (e-mail: p.ghamisi@ gmail.com).}}

\maketitle

\begin{abstract}
The use of deep learning for water extraction requires precise pixel-level labels. However, it is very difficult to label high-resolution remote sensing images at the pixel level. Therefore, we study how to utilize point labels to extract water bodies and propose a novel method called the neighbor feature aggregation network (NFANet). Compared with pixel-level labels, point labels are much easier to obtain, but they will lose much information. In this paper, we take advantage of the similarity between the adjacent pixels of a local water-body, and propose a neighbor sampler to resample remote sensing images. Then, the sampled images are sent to the network for feature aggregation. In addition, we use an improved recursive training algorithm to further improve the extraction accuracy, making the water boundary more natural. Furthermore, our method utilizes neighboring features instead of global or local features to learn more representative features. The experimental results show that the proposed NFANet method not only outperforms other studied weakly supervised approaches, but also obtains similar results as the state-of-the-art ones.
\end{abstract}


%
\IEEEpeerreviewmaketitle

\section{Introduction}

\IEEEPARstart{W}{ater} body extraction from high-resolution remote sensing images is an important research topic in the field of remote sensing applications. Although the traditional algorithms have made some progress in water-body extraction, there are still problems such as cumbersome manual feature extraction and insufficient extraction accuracy. At the same time, with the rapid advancement of remote sensing technology, high-resolution remote sensing images contain increasingly detailed texture information and edge structure information. With traditional remote sensing water-body extraction methods, it is difficult to make full use of the semantically rich high-resolution remote sensing image information, and also difficult to meet the growing demand of remote sensing applications.

\begin{figure}
	\centering
	\includegraphics[width=1\linewidth]{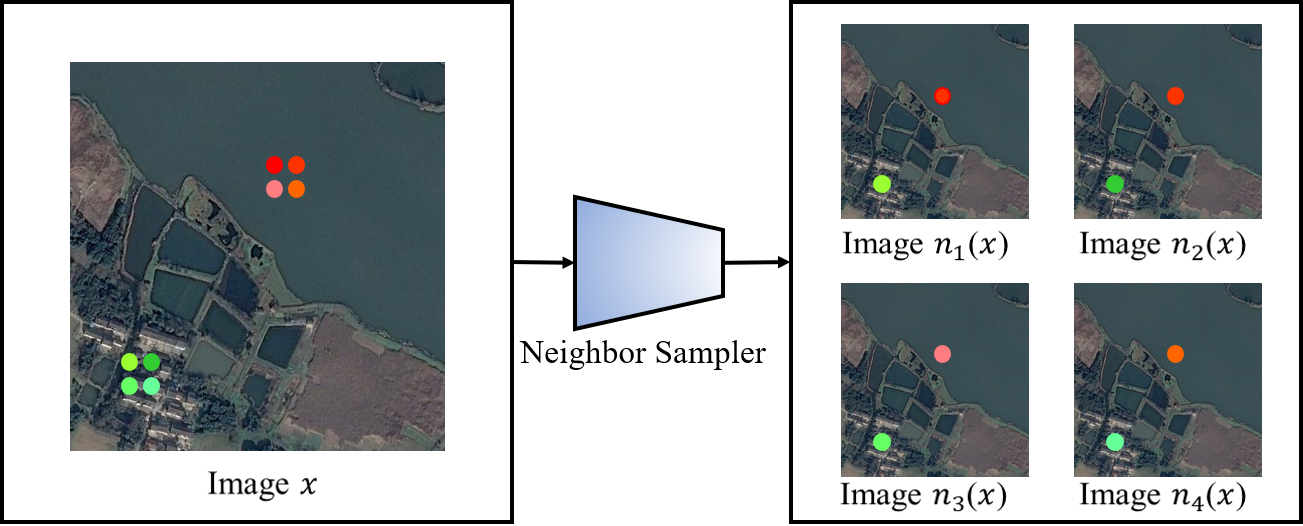}
	\caption{Use of the neighbor sampler to sample the input image, where the left is the original image, and the right is the neighbor image group. The dots of different colors indicate the pixel mapping relationship.}
	\label{fig:1}
\end{figure}

In recent years, deep learning has become an emerging research hot spot in the field of artificial intelligence \cite{8277160,9119167,8792386,9291440}. The rapid development of deep learning technology and the improvement of computer hardware performance have enabled deep learning, especially convolutional neural network (CNN) \cite{9167434,8580422,8827587}, to be successfully applied to many important tasks, such as image classification \cite{9018274,8502139}, target detection \cite{8879668,8796403}, and semantic segmentation \cite{8736792,8576646}. In addition, due to its excellent performance, deep learning has been widely used in the remote sensing image fields, such as remote sensing image classification \cite{9271913,9345971,9431581,9347809,9366971}, change detection \cite{9291461,9036876}, and ground object extraction \cite{9324236,9094008}. Hong \emph{et al.} \cite{9174822} proposed a variety of fusion architectures to solve the special case of multi-modal learning and cross-modality learning that is widely used in remote sensing image classification applications. Hong \emph{et al.} \cite{9170817} established a method that combined graph convolutional network (GCN) and CNN to fuse different hyperspectral features to improve the performance of hyperspectral classification.

Deep learning is also used for water extraction, and the deep learning-based water extraction method can significantly improve the extraction accuracy of traditional techniques. Zhang \emph{et al.} \cite{9118970} proposed a cascaded fully-convolutional network to improve the performance of water-body detection, and introduced fully-convolutional conditional random fields (CRF) to realize automatic learning of Gaussian kernels in CRF. Li \emph{et al.} \cite{9360447} proposed a dense local feature compression network. Each layer of this network was densely connected to receive all its previous feature maps. In addition, the network included a local feature compression (LFC) module to integrate spatial and spectral information. He \emph{et al.} \cite{9444849} applied UNet and attention-UNet to Landsat-8 images for glacial lakes extraction. Specifically, the self-attention module was added to the skip connection to strengthen the extraction of the network features of the glacial lake. Wang \emph{et al.} \cite{8989826} combined the multi-scale convolutional network with Google Earth Engine and proposed an offline training and online prediction method to monitor urban water. In order to improve the ability to extract the boundaries of water bodies, Miao \emph{et al.} \cite{8286914} proposed a new loss function called edges weighting loss, which calculated the Euclidean distance to obtain a higher weight for the boundary.

In general, the success of deep learning for feature extraction is highly dependent on the availability of sufficient pixel-level labels for training. However, high-resolution remote sensing images are large in both scale and data volume, which makes pixel-level labeling extremely laborious. Pixel-level annotation usually requires a lot of time and labor costs, and professional knowledge to accurately mark uncertain boundaries between different classes of interest, which hinders the development of high-resolution remote sensing image feature extraction to a certain extent. Training algorithm models using weak labels has received increasing attention in the field of computer vision \cite{9353394,7971941,8360132}. Compared with fully-supervised semantic segmentation, weak-supervised learning  \cite{9143431,9355009,9190782} does not require pixel-level labels, and has the characteristics of fast labeling and low time cost. However, the use of weak annotations makes the supervision information seriously insufficient, and thus the key information, such as shape, texture, and edges, is usually lost, which makes it difficult to extract water from high-resolution remote sensing images with complex scenes.

In this paper, we propose a new solution to the above problems. Unlike other natural objects, water-bodies are usually liquid, whose colors and textures are very similar. Therefore, there is a high degree of similarity between neighbor pixels in water-bodies, which makes the inherent difference between the neighbor pixels of the water-body generally smaller than those of the non-water-body. Thus, we attempt to map the neighbor pixels of remote sensing images to the same location in space, and then extract neighbor features from multiple neighbor pixels. Next, we use the neighbor features to jointly decide whether the pixel at this location belongs to the water-body. Based on the above motivation, we propose the neighbor feature aggregation network (NFANet) to make full use of this property. Specifically, we utilize a sampling method called neighbor sampler to generate a set of neighbor images from high-resolution remote sensing images. The neighbor pixels of the original image are allocated separately to each neighbor image, so that the pixel values of any two neighbor images at the same position are similar, but the pixel values are different. On the whole, neighbor image groups have similar but different characteristics (see Fig. 
\ref{fig:1}). Once the neighbor images are generated, we use an end-to-end model to perform feature extraction on each image of the neighbor image groups, and aggregate the features by using a feature aggregation module. Compared with other methods that only use the local or global information from the image, neighbor feature aggregation effectively utilizes the neighbor information, and so more representative features can be learned. Finally, we used recursive training to further refine the water-body boundary, making the water-body edge smoother and more natural. Our experiments demonstrate that the proposed NFANet can utilize neighbor information and achieve good results, that are comparable to the results of the current state-of-the-art fully-supervised methods. 

The main contributions of this paper are summarized as follows.

\begin{enumerate}
{\item	We propose a new sampling method to resample high-resolution remote sensing images and generate neighbor image groups in order to extract neighbor information, which helps to distinguish between water-bodies and land boundaries.}

{\item	We develop a weakly supervised deep learning algorithm model called NFANet. Compared with other methods that only use local features or global features, the proposed approach enables using neighbor information to be used to learn more representative characteristics.}

{\item	We propose a recursive training strategy, which can further refine the boundary of the water-body, making the boundary smooth and closer to the ground truth.}
\end{enumerate}

This paper is organized as follows. Section II introduces some related works and the main contribution of our method. Section III discusses the proposed NFANet and related modules in detail. In section IV, a comprehensive experiment is carried out on the proposed method on the water-body data-set, and a comparison to other state-of-the-art methods is given. Section V summarizes the conclusions of this paper.

\begin{figure*}
	\centering
	\includegraphics[width=1\linewidth]{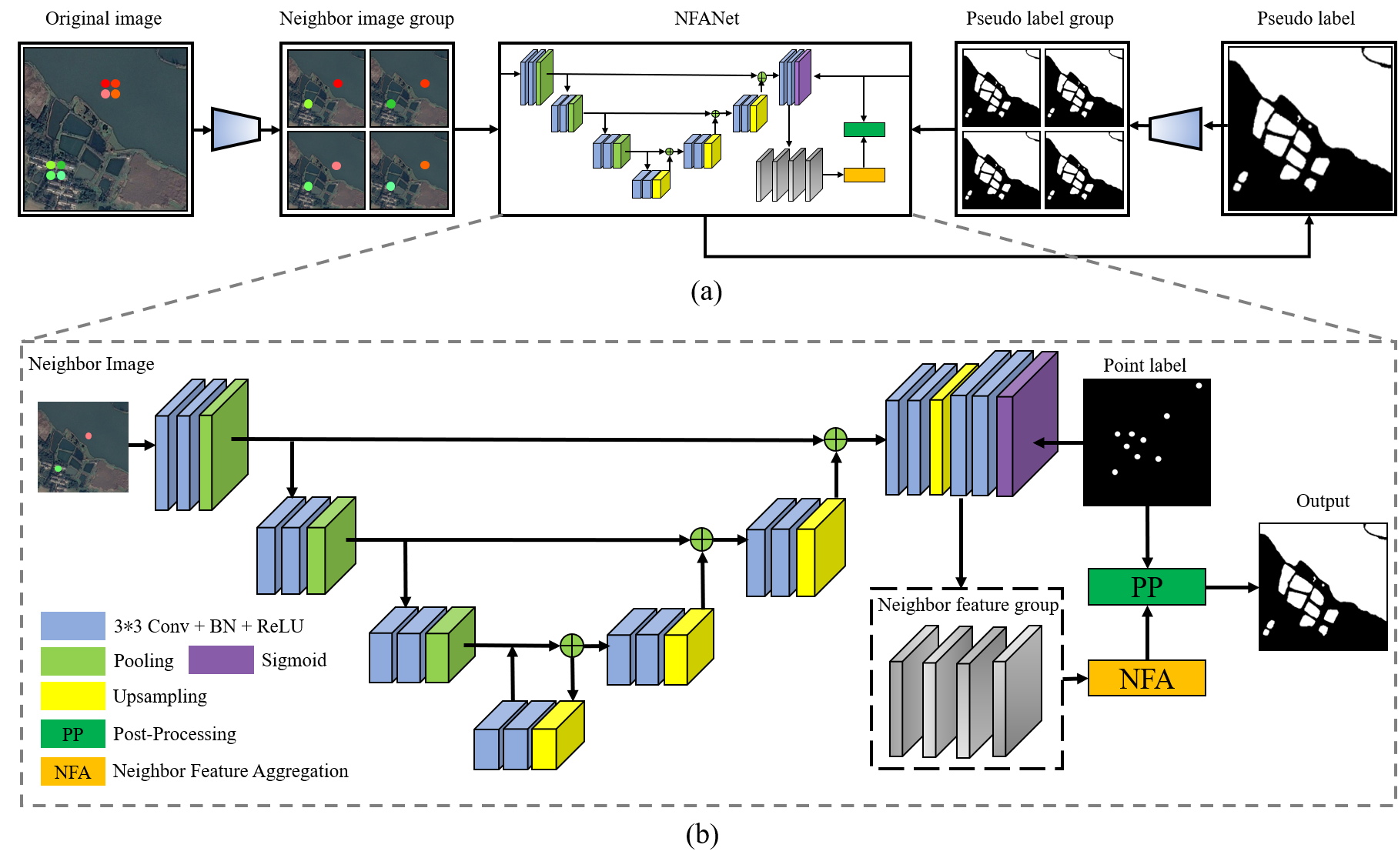}
	\caption{The proposed weakly supervised water-body extraction framework; (a) represents the recursive training process, and (b) is the proposed NFANet.}
	\label{fig:2}
\end{figure*}

\section{Related Works and Contributions}

\subsection{Related Works}
In the past few years, a large number of approaches have been proposed to solve the problem of water-body extraction from remote sensing images. Generally, these methods can be divided into two categories: methods based on manual features and methods based on deep learning.
	
\subsubsection{Methods based on manual features}
The key problem of water-body extraction is how to determine the difference between the water-body and other natural features. For this reason, methods based on manual features usually hope to find one or more distinctive features that can distinguish water bodies and non-water bodies. Gao \emph{et al}. \cite{Bo1995NDWI} proposed the normalized differential water index method (NDWI), which uses a specific band of remote sensing images for normalized difference processing. This method can highlight the water information in the image and weaken the influence of non-water factors. Trias-Sanz \emph{et al.} \cite{1704962} used a hierarchical segmentation method combining color space transformation and texture features to segment the lake. Barton \emph{et al}. \cite{1989Monitoring} used the brightness temperature extracted by AVHRR channel 4 to identify the water-body and monitor flooding.

\subsubsection{Methods based on deep learning}
Although there has been some progress in the task of water-body extraction, manual features still have problems, such as cumbersome design and limited extraction accuracy. Recently, with the powerful generalization ability of deep learning, related models have been widely used in the field of water extraction. In order to balance the background semantics and edge details, Zhang \emph{et al.} \cite{9151932} proposed a network based on R-CNN called Mask-R-CNN, which combines semantic segmentation and target detection at a decision-level, and improved the segmentation accuracy of water, pools, roads, and buildings. Li \emph{et al.} \cite{8883176} argued that a single-scale model may lead to inaccurate extraction of water bodies. They combined DeepLab v3+ with multi-scale joint prediction and used fully connected CRF to optimize the water-body boundary. Wang \emph{et al.} \cite{8620300} used a model accuracy variation degree algorithm and a sample quality analysis algorithm to reduce the impact of incorrect water labeling on the model, thereby improving prediction accuracy. Cui \emph{et al.} \cite{9269403} proposed a multi-scale feature extraction module and an adaptive feature fusion module and applied them to sea-land segmentation. Chu \emph{et al.} \cite{8900625} constructed Res-UNet to handle complex sea-land segmentation scenes, and then used fully connected CRF and morphological operations to further refine the water-body boundary. Dong \emph{et al.} \cite{8898367} introduced the spatial position and feature space of adjacent pixels into the loss function, and proposed a sub-neighbor system constraint. Feng \emph{et al.}. \cite{8573826} combined the super-pixels-based CRF and prediction results of UNet to enhance water extraction.

\subsection{Weakly supervised learning}
Fully-supervised learning can extract the target object in a remote sensing image very accurately, but usually requires the use of pixel-level labels, which consumes a lot of manpower and financial resources, and needs a certain degree of professional knowledge for labeling. Some researchers have tried to use traditional methods combined with deep learning to solve weak supervision problems. Fu \emph{et al}. \cite{2018WSF} combined super-pixels and a local map to obtain rough pseudo-labels to train a water extraction model. Chen \emph{et al.} \cite{2020SPMF} combined super-pixel pooling with multi-scale feature fusion to detect buildings. Other researchers attempted to obtain better results by using the extraction capabilities of the neural network itself. Wang \emph{et al.} \cite{rs12020207} learned from the principle of class activation maps (CAM) \cite{2016Learning} and extracts feature maps from UNet \cite{2015U} for hard-threshold processing to obtain segmentation predictions. These methods have achieved promising results in the field of weak-supervised learning, but do not consider the characteristics of the image itself.

\subsection{The novelty of the proposed method}
The proposed method belongs to the category of weakly supervised learning methods. Due to the similarity of neighbor pixels in local water bodies, the potential features between neighbor pixels in local water-bodies are more alike than those in non-water-bodies. Based on this characteristic, a neighbor sampler is proposed to resample remote sensing images and use a feature aggregation module to obtain rough results. The proposed method uses the features of neighbor pixels instead of only the global features or local features of the image, and so more features can be learned, which is more representative. Compared with methods such as Unet-based CAM (U-CAM) \cite{rs12020207}, our approach exhibits better performance and is comparable to the visual results of the fully-supervised method. In addition, it is inspired by the fact that when applying the resulting model over the training set, the outputs of the network can significantly better capture the shape of objects than that of just pseudo-labels \cite{8099664}. We have observed through experiments that when the training set is input into the network again, the obtained network output will become smoother than the coarse-grained pseudo-label, which improves the accuracy of the prediction result to a certain extent.

\section{Proposed Method}

Fig. \ref{fig:2} illustrates the proposed weakly supervised water extraction framework. Fig. \ref{fig:2}a shows the entire recursive training process, which is described in subsection C below. The acquisition of pseudo-labels is shown in Fig. \ref{fig:2}b. We input neighbor images into the network and use point labels for supervision to obtain neighbor features. The feature aggregation module is then used for feature aggregation. Finally, post-processing is performed to obtain pseudo-labels. We describe the details of each of the above steps in the following sections.

\begin{figure}
	\centering
	\includegraphics[width=1\linewidth]{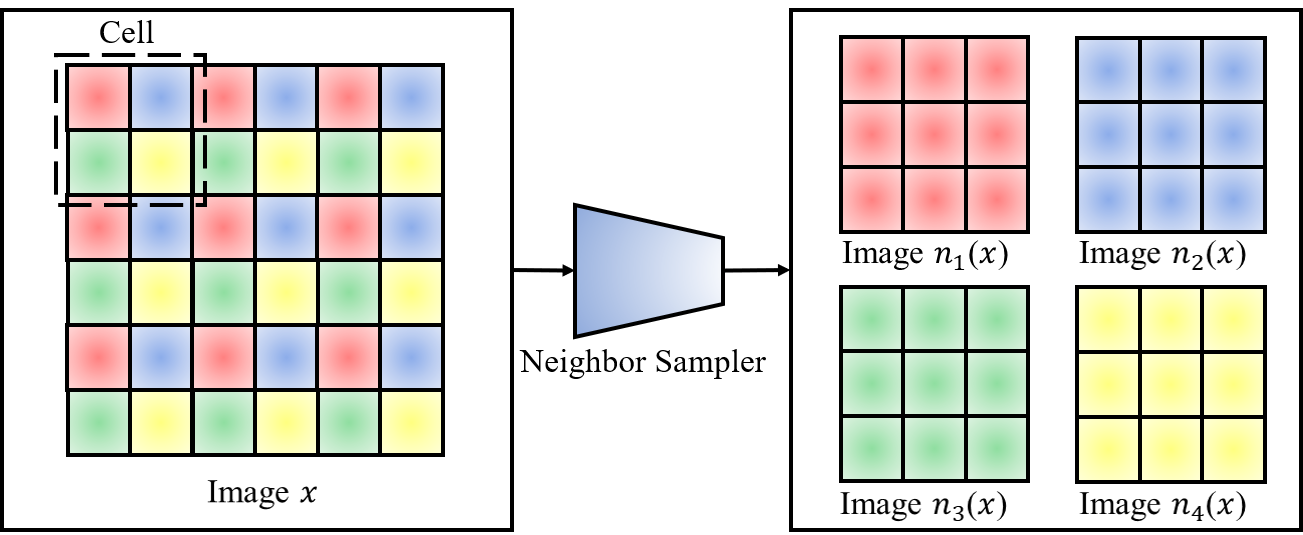}
	\caption{Proposed neighbor sampler. When k is set to 2, a cell contains neighbor pixels of \begin{math} 2 \times 2 \end{math} size and is reallocated to a neighbor image group. Best viewed in color.}
	\label{fig:3}
\end{figure}

\subsection{Neighbor Sampler}
First, we introduce a neighbor sampler to obtain a neighbor images group \begin{math}
	(n_1\left(x\right),n_2\left(x\right),\ldots{,n}_L\left(x\right)) \end{math} from a single optical remote sensing image \begin{math} x \end{math}. \begin{math} L \end{math} represents the number of neighbor images. Fig. \ref{fig:3} shows the schematic diagram of generating a group of neighbor images using the neighbor sampler. Let us assume that the width, height, and channel of the input image \begin{math} x \end{math} are \begin{math} W,H \end{math}, and \begin{math} C \end{math}, respectively. The implementation of the neighbor sampler \begin{math} N=(n_1,n_2,\ldots{,n}_L) \end{math} is described as follows:

\begin{enumerate}
{\item
The image \begin{math} x \end{math} is divided into \begin{math} \frac{W}{K}\times\frac{H}{K} \end{math} cells, where the size of each cell is \begin{math} K\times K\times C \end{math}. We experimentally set \begin{math} K \end{math} to 2 and, therefore, \begin{math} L=K\times K=4 \end{math}.}

{\item
For the \begin{math} i-th \end{math} row and \begin{math} j-th \end{math} column of the cell, the pixels in the adjacent positions of each cell are selected in the order from top to bottom and from left to right, which are regarded as the \begin{math} (i,j)-th \end{math} elements of \begin{math} N=(n_1\left(x\right),n_2\left(x\right),\ldots{,n}_l\left(x\right)) \end{math}. When \begin{math} K \end{math} is set to 2, the pixels at the upper left, upper right, lower left, and lower right adjacent positions are selected, respectively.}

{\item
For all \begin{math} \frac{W}{K}\times\frac{H}{K} \end{math} cells being divided in step 1), step 2) will be repeated until all the cells are resampled, and a neighbor sampler \begin{math} N=(n_1,n_2,\ldots{,n}_L) \end{math} is generated. Given an optical remote sensing image \begin{math} x \end{math}, a neighbor images group \begin{math} (n_1\left(x\right),n_2\left(x\right),\ldots{,n}_L\left(x\right)) \end{math} is generated, where the size of each neighbor image is \begin{math} \frac{W}{K}\times\frac{H}{K}\times C \end{math}.}

\end{enumerate}

In this way, the neighbor image dataset can be generated from the original dataset. Neighbor images are similar but not identical, because for any two neighbor images, the \begin{math} (i,j)-th \end{math} pixel comes from the neighboring location of the original remote sensing image.

\subsection{Feature aggregation module and post-processing}

\begin{figure}
	\centering
	\includegraphics[width=1\linewidth]{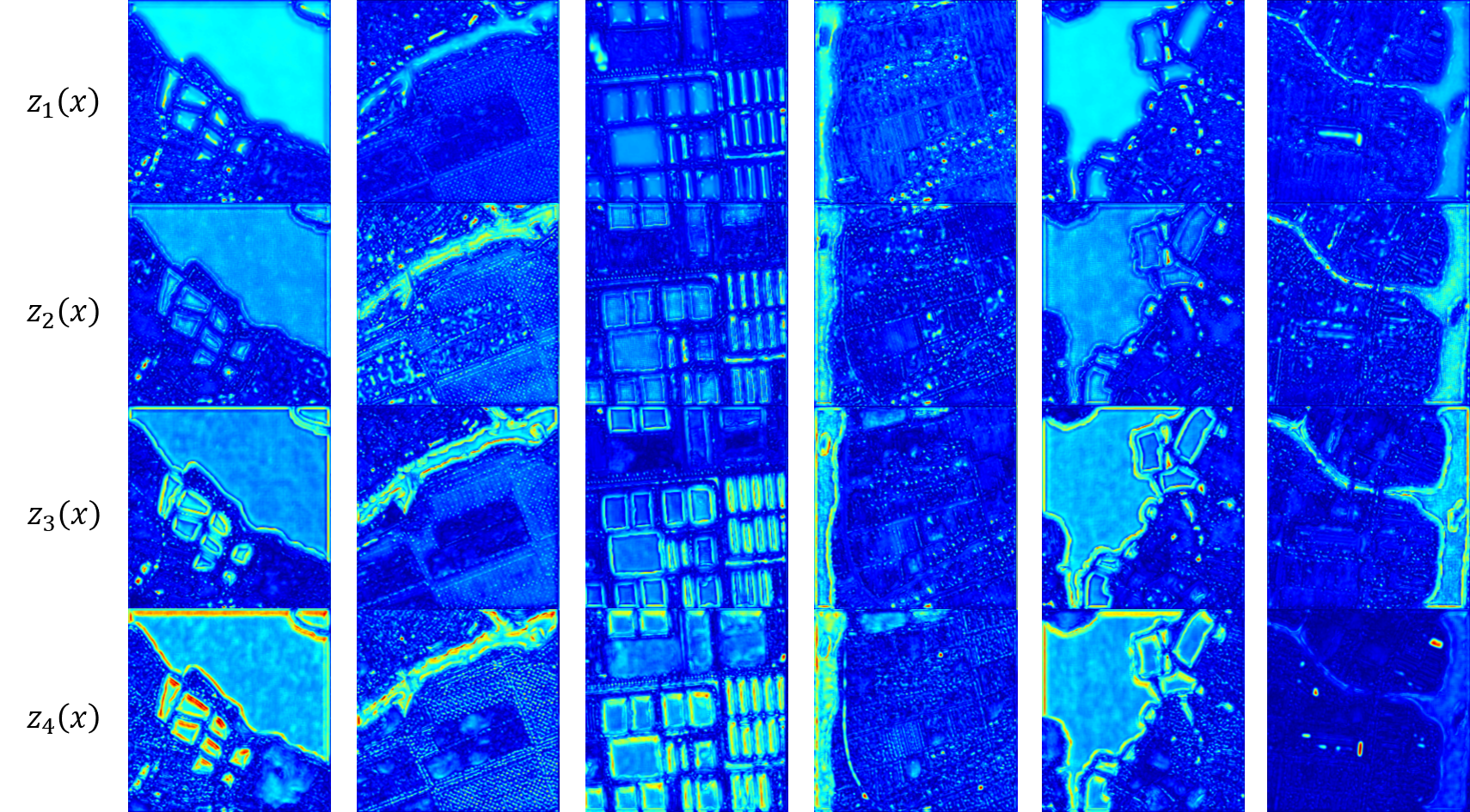}
	\caption{Visualization results of neighbor feature groups after CMax pooling. Columns represent the same neighbor feature group.}
	\label{fig:4}
\end{figure}

We input the neighbor images group into an end-to-end network to extract features, and obtain the corresponding neighbor feature group \begin{math} \left(f_1\left(X\right),f_2\left(X\right),\ldots{,f}_L\left(X\right)\right)\in\ \mathbb{R}^{H\times W\times C\times L} \end{math}, where \begin{math} f_l\left(X\right)\ \in\ \mathbb{R}^{H\times W\times C} \end{math} represents the feature maps extracted from the \begin{math} l-th \end{math} 
image in the neighbor images group. We use the encoder-decoder structure as the feature extraction network. Specifically, the feature maps are extracted from the penultimate convolutional layer. The network structure is shown in Fig. \ref{fig:2}b. It is worth noting that the network is replaceable (in the experimental part, a variety of network structures are used for feature extraction). CMax pooling is adopted to reduce the number of channels of each neighbor feature to one. CMax pooling is defined mathematically in detail as follows: Given a three-dimensional feature maps tensor group  \begin{math} F=\left(f_1\left(x\right),f_2\left(x\right),\ldots{,f}_L\left(x\right)\right)\in\ \mathbb{R}^{H\times W\times C\times L} \end{math}, the operation of CMax pooling is as follows:
\begin{normalsize} 
	\begin{equation}
		\begin{aligned}
			z_{i,j,l}(x)=\mathop{max}(f_{i,j,1,l}(x),f_{i,j,2,l}(x),\ldots,f_{i,j,3,l}(x)),\\ i=1,2,\ldots,H, j=1,2,\ldots,w, l=1,2,\ldots,L.
		\end{aligned}
	\end{equation}
\end{normalsize} 
\begin{figure*}
	\centering
	\includegraphics[width=1\linewidth]{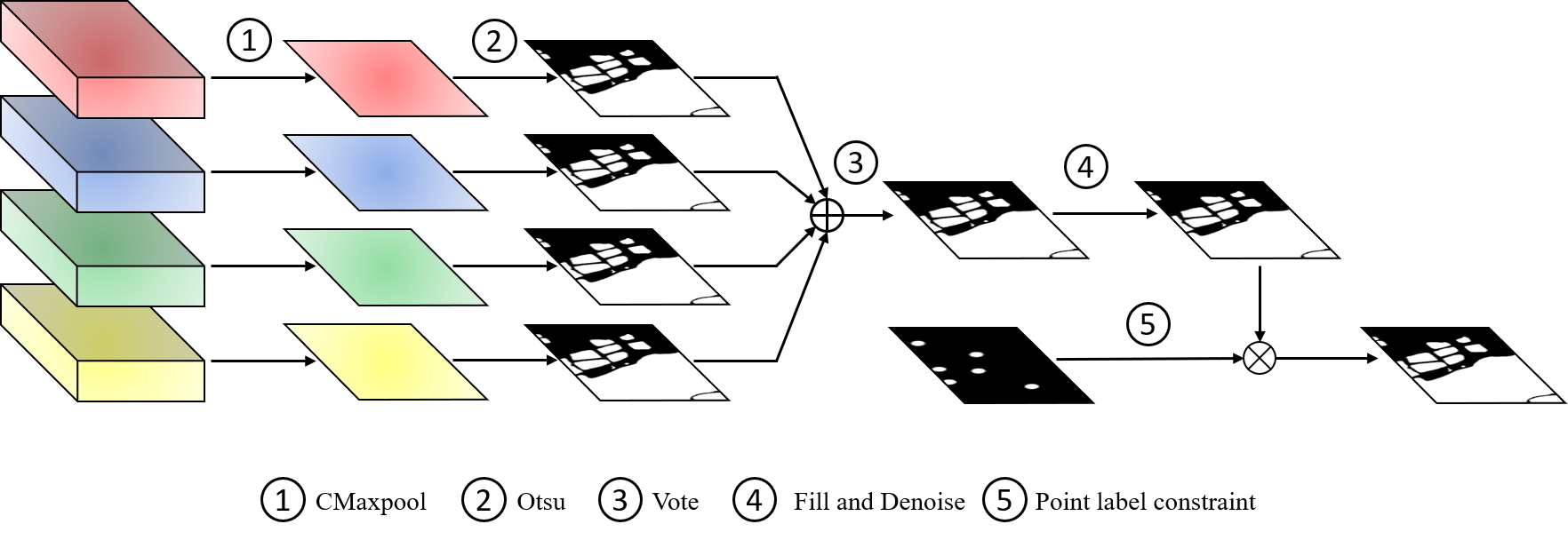}
	\caption{Feature aggregation module and post-processing.}
	\label{fig:5}
\end{figure*}
As a result, the feature maps group  \begin{math} Z=\left(z_1\left(x\right),z_2\left(x\right),\ldots,z_L\left(x\right)\right)\ \in\mathbb{R}^{H\times W\times L} \end{math} is obtained. Fig. \ref{fig:4} shows the visualization results of the feature map group \begin{math} Z \end{math}. It can be seen that different neighbor images of the same remote sensing image have a certain degree of attention to the water-body. Some features pay more attention to the boundary between a water-body and non-water-body, and some features pay more attention to texture areas. 

Next, the OTSU algorithm is used to binarize each feature in \begin{math} Z \end{math} to obtain the result \begin{math} O=\left(o_1\left(x\right),o_2\left(x\right),\ldots,o_L\left(x\right)\right)\ \in\mathbb{R}^{H\times W\times L} \end{math}. The formula is as follows:
\begin{equation}
	o_l=Otsu\left(z_l\right), l=1,2,\ldots,L. 
\end{equation}
Finally, we vote for all binarized neighbor features of the neighbor feature group to obtain the aggregated result \begin{math} V \end{math}.  The specific formula is calculated as:
\begin{equation}
	V_{i,j}= \begin{cases}
		1,\sum^L_{l=1}o_{i,j,l}\geq \frac{2}{L}\\
		0,\sum^L_{l=1}o_{i,j,l}< \frac{2}{L}
	\end{cases},
\end{equation}
To summarize, the mathematical definition of the feature aggregation module is detailed as follows:
\begin{equation}
	V=Vote\left(Otsu\left(CMax\left(F\right)\right)\right),
\end{equation}
where \begin{math} F\in\ \mathbb{R}^{H\times W\times C\times L} \end{math} represents the neighbor features group and \begin{math} V\in\mathbb{R}^{H\times W} \end{math} is the output. Next, the aggregated result \begin{math} V \end{math} is input into the post-processing module. Fig. \ref{fig:5} shows the feature aggregation module and post-processing module. The specific operations include filling small holes in the closed area by using area filling and removing noise by using morphological operations. Then we apply a point label constraint to the processed results. The point label constraint is expressed as follows:
\begin{equation}
	q= \begin{cases}
		1,\ p\subset Q\\
		0,\ p\not\subset Q
	\end{cases}.
\end{equation}
where \begin{math} Q\in\mathbb{R}^{H\times W} \end{math} represents any independent water-body area in \begin{math} V \end{math}, \begin{math} q \end{math} is any pixel in \begin{math} Q \end{math}, and \begin{math} p \end{math} represents the point labels. If the area in the result contains point labels, the entire area is retained, otherwise it is not retained; The generated results are used as pseudo-labels and input into recursive training as supervision information.

\subsection{Recursive training}

There exist several recursive training approaches in the field of deep learning \cite{8394987,Wang_2018_CVPR}. The work in \cite{Shen_2019_CVPR} uses a semantic segmentation and target detection branch for cyclic guidance to improve the quality of pseudo-labels.
In \cite{Xie_2020_CVPR}, data noise and model noise are used for recursive training to improve the generalization ability of the model. Unlike the above methods, we embed the neighbor sampler into the recursive training so that the network can learn neighbor features ( the flowchart is shown in Fig. \ref{fig:2}a ). The specific steps of our method are as follows:\par

\begin{enumerate}
	{\item[a.] Input the images into the neighbor sampler to generate neighbor image groups, and input these neighbor images with point labels into the model for training.}
	
	{\item[b.] Input the neighbor images into the model again, and use the feature aggregation module to generate pseudo-labels as a replacement for point labels.	}
	
	{\item[c.] Input pseudo-labels into the neighbor sampler to generate pseudo-labels groups, and input the neighbor images groups with pseudo-labels groups into the model for training. }
	
	{\item[d.] Input the neighbor images groups into the model again for prediction, and generate results groups (average the group results to obtain a new pseudo-label) as the new pseudo-labels to replace the previous set of labels. }
	
	{\item[e.] Go to step c. }
\end{enumerate}

\section{Experimental Results And Discussion}

\begin{figure*}
	\centering
	\includegraphics[width=1\linewidth]{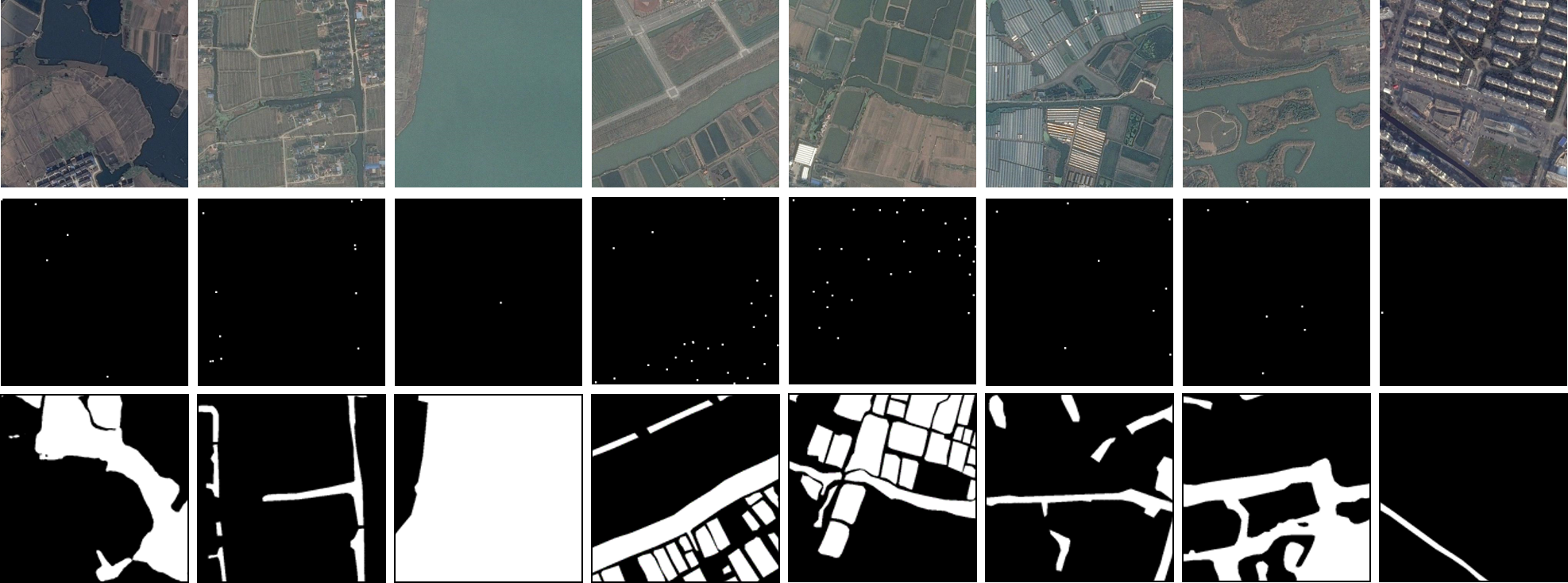}
	\caption{Water-body extraction data-set. The first line represents the original images, the second line represents the point labels, and the third line represents ground truth.}
	\label{fig:6}
\end{figure*}

\subsection{Datasets and Evaluation}
To verify the effectiveness of the proposed method, we applied the proposed method to high-resolution visible spectrum images for water extraction. This water-body dataset comes from the Gaofen Challenge \cite{9185011}, which contains RGB pan-sharpened images with a resolution of 0.5 m and does not contain infrared bands or digital elevation models. All images are taken from Wuhan and Suzhou, China, mainly in rural areas supplemented by urban areas. The positive labels in the dataset include rivers, reservoirs, rice fields, ditches, ponds, and lakes, while all other non-water pixels are considered negative. The data-set was cropped into 1000 images with the size of \begin{math} 492\times492 \end{math} without any overlap. The original data-set only contained pixel-level labels, so we re-annotated the data-set. Specifically, the data-set was annotated by two professionals and checked by an expert in the field of remote sensing image processing. The rule was that each independent water-body was randomly labeled with a point label of size \begin{math} 5\times5 \end{math}, which simulated the labeling behavior of the labelers in a real scene. We utilized the pixel-level labels of the original data-set as ground truth. The data sets are shown in Fig. \ref{fig:6}.
In the experiment, the weak supervision models adopt the point labels as the initial supervision information, while the full supervision models use pixel-level labels. Because the remote sensing image segmentation/classification evaluation index of overall accuracy or kappa coefficient cannot effectively describe the real structure of image segmentation geometry, we choose the fgIoU (foreground IoU), bgIoU (background IoU), mIoU (mean IoU), fgDice (foreground Dice), bgDice (background Dice), and mDice (mean Dice) to comprehensively evaluate the results. For each model, we performed five independent runs to calculate these evaluation indicators and standard deviations.

\begin{table*}
	\setlength{\abovecaptionskip}{0.pt}
	\setlength{\belowcaptionskip}{-0.em}
	\renewcommand\tabcolsep{3.5pt} 
	\centering
	\scriptsize
	\caption{Comparison of Water Extraction Results (\%) and Full Supervision Approaches}
	\begin{tabular}{ccccccccccccccc}
		\hline
		\hline 
		\multirow{2}{*}{Method} &\multirow{2}{*}{Sup.} & \multicolumn{6}{c}{Validation} & \multicolumn{6}{c}{Test}  \\
		\cmidrule(r){3-8} \cmidrule(r){9-14}
		& &bgIoU &fgIoU  &mIoU &bgDice &fgDice &mDice &bgIoU &fgIoU  &mIoU &bgDice &fgDice &mDice \\
		\hline 
		FCN &full  &91.2$\pm$0.4 &66.9$\pm$1.0 &79.1$\pm$0.5 &95.4$\pm$0.2 &80.2$\pm$0.7 &87.8$\pm$0.4 &91.0$\pm$0.6 &66.4$\pm$0.5 &78.7$\pm$0.4 &95.3$\pm$0.3 &79.8$\pm$0.3 &87.6$\pm$0.2 \\
		
		FCN(Ours) &weak  &89.6$\pm$0.2 &61.5$\pm$0.6 &75.5$\pm$0.4 &94.5$\pm$0.1 &76.1$\pm$0.5 &85.3$\pm$0.3 &89.3$\pm$0.4 &60.7$\pm$1.0 &75.0$\pm$0.7 &94.4$\pm$0.2 &75.5$\pm$0.7 &85.0$\pm$0.5	\\
		
		UNet &full  &92.7$\pm$0.2	&73.3$\pm$0.4	&83.0$\pm$0.3	&96.2$\pm$0.1	&84.6$\pm$0.3	&90.4$\pm$0.2	&92.6$\pm$0.1	&73.0$\pm$0.3	&82.8$\pm$0.2	&96.2$\pm$0.1	&84.4$\pm$0.2	&90.3$\pm$0.1  \\
		
		UNet(Ours) &weak  &91.4$\pm$0.4 &67.8$\pm$1.3 &79.6$\pm$0.8 &95.5$\pm$0.2 &80.8$\pm$0.9 &88.2$\pm$0.5 &91.1$\pm$0.3 &67.0$\pm$0.7 &79.1$\pm$0.5 &95.3$\pm$0.2 &80.3$\pm$0.5 &87.8$\pm$0.3	\\
		
		Res-UNet &full &92.7$\pm$0.2 &73.3$\pm$0.6	&83.0$\pm$0.4	&96.2$\pm$0.1	&84.6$\pm$0.4	&90.4$\pm$0.2	&92.6$\pm$0.1	&72.9$\pm$0.4	&82.7$\pm$0.2	&96.1$\pm$0.1	&84.3$\pm$0.2	&90.2$\pm$0.2	\\

		Res-UNet(Ours) &weak  &91.2$\pm$0.3 &67.4$\pm$0.7 &79.3$\pm$0.5 &95.4$\pm$0.2 &80.5$\pm$0.5 &88.0$\pm$0.3 &91.0$\pm$0.2 &67.0$\pm$0.4 &78.9$\pm$0.4 &95.3$\pm$0.1 &80.1$\pm$0.4 &87.7$\pm$0.2	\\
		
		NestedUNet &full  &93.0$\pm$0.4	&74.2$\pm$0.4	&83.6$\pm$0.3	&96.4$\pm$0.2	&85.2$\pm$0.2	&90.8$\pm$0.2 &92.8$\pm$0.3	&73.3$\pm$0.6	&83.1$\pm$0.2	&96.3$\pm$0.2	&84.6$\pm$0.4	&90.4$\pm$0.2 \\  
		
		NestedUNet(Ours) &weak  &\textbf{91.7}$\pm$\textbf{0.2} &\textbf{69.0}$\pm$\textbf{0.6} &\textbf{80.4}$\pm$\textbf{0.4} &\textbf{95.7}$\pm$\textbf{0.1} &\textbf{81.7}$\pm$\textbf{0.4} &\textbf{88.7}$\pm$\textbf{0.3} &\textbf{91.6}$\pm$\textbf{0.2} &\textbf{68.8}$\pm$\textbf{0.6} &\textbf{80.2}$\pm$\textbf{0.4} &\textbf{95.6}$\pm$\textbf{0.1} &\textbf{81.5}$\pm$\textbf{0.4} &\textbf{88.6}$\pm$\textbf{0.3}	\\
		
		D-LinkNet &full  &92.3$\pm$0.2	&72.3$\pm$0.7	&82.3$\pm$0.5	&96.0$\pm$0.1	&83.9$\pm$0.5	&90.0$\pm$0.3	&92.0$\pm$0.2	&71.1$\pm$0.3	&81.6$\pm$0.2	&95.9$\pm$0.1	&83.1$\pm$0.2	&89.5$\pm$0.2
		\\
		
		D-LinkNet(Ours) &weak  &91.1$\pm$0.2 &67.1$\pm$0.4 &79.1$\pm$0.3 &95.3$\pm$0.1 &80.3$\pm$0.3 &87.8$\pm$0.2 &90.8$\pm$0.2 &66.1$\pm$0.6 &78.5$\pm$0.4 &95.2$\pm$0.1 &79.6$\pm$0.4 &87.4$\pm$0.3	\\
		
		DeepLab V3+ &full  &93.3$\pm$0.2 &75.3$\pm$0.4	&84.3$\pm$0.3	&96.5$\pm$0.1	&85.9$\pm$0.2	&91.2$\pm$0.2	&92.9$\pm$0.3	&74.3$\pm$0.7	&83.6$\pm$0.5	&96.3$\pm$0.1	&85.3$\pm$0.4	&90.8$\pm$0.3
		\\ 
		DeepLab V3+(Ours) &weak  &91.5$\pm$0.1  &68.5$\pm$0.3	&80.0$\pm$0.2	&95.6$\pm$0.1	&81.3$\pm$0.2	&88.4$\pm$0.1	&91.2$\pm$0.2	&67.6$\pm$0.6	&79.4$\pm$0.4	&95.4$\pm$0.1	&80.6$\pm$0.5	&88.0$\pm$0.3		\\
		
		MFDeepLab V3+ &full &93.4$\pm$0.2	&75.6$\pm$0.4	&84.5$\pm$0.3	&96.6$\pm$0.1	&86.1$\pm$0.3	&91.3$\pm$0.2	&93.0$\pm$0.2	&74.5$\pm$0.5	&83.7$\pm$0.3	&96.4$\pm$0.1	&85.4$\pm$0.3	&90.9$\pm$0.2
		\\
		
		MFDeepLab V3+(Ours) &weak  &91.6$\pm$0.1 &68.7$\pm$0.3 &80.1$\pm$0.2 &95.6$\pm$0.0 &81.4$\pm$0.2 &88.5$\pm$0.1 &91.2$\pm$0.2 &67.8$\pm$0.6 &79.5$\pm$0.4 &95.4$\pm$0.1 &80.8$\pm$0.4 &88.1$\pm$0.3	\\
		
		SANet &full &93.0$\pm$0.1 &74.2$\pm$0.2 &83.6$\pm$0.2 &96.4$\pm$0.1 &85.2$\pm$0.1 &90.8$\pm$0.1 &92.7$\pm$0.1 &73.3$\pm$0.2 &83.0$\pm$0.2 &96.2$\pm$0.1 &84.6$\pm$0.2 &90.4$\pm$0.1 \\
		
		SANet(Ours) &weak  &91.4$\pm$0.2 &68.1$\pm$0.3 &79.8$\pm$0.3 &95.5$\pm$0.1 &81.0$\pm$0.2 &88.3$\pm$0.2 &91.1$\pm$0.2 &67.2$\pm$0.6 &79.2$\pm$0.4 &95.3$\pm$0.1 &80.4$\pm$0.4 &87.9$\pm$0.3	\\
		
		SNS-CNN &full &93.2$\pm$0.2	&74.6$\pm$0.2	&83.9$\pm$0.2	&96.3$\pm$0.5	&85.4$\pm$0.1	&91.0$\pm$0.1	&92.9$\pm$0.2	&73.8$\pm$0.4	&83.3$\pm$0.2	&96.3$\pm$0.1	&84.9$\pm$0.3	&90.6$\pm$0.1 	\\
		
		SNS-CNN(Ours) &weak  &91.6$\pm$0.1 &68.6$\pm$0.4 &80.1$\pm$0.3 &95.6$\pm$0.1 &81.4$\pm$0.3 &88.5$\pm$0.2 &91.4$\pm$0.3 &68.2$\pm$0.7 &79.7$\pm$0.5 &95.5$\pm$0.2 &81.1$\pm$0.5 &88.3$\pm$0.3	\\
		
		DeepUNet &full  &92.9$\pm$0.1	&73.6$\pm$0.3	&83.2$\pm$0.2	&96.3$\pm$0.1	&84.8$\pm$0.2	&90.5$\pm$0.1	&92.6$\pm$0.1	&72.9$\pm$0.3	&82.8$\pm$0.2	&96.2$\pm$0.1	&84.3$\pm$0.2	&90.2$\pm$0.1	\\
		
		DeepUNet(Ours) &weak  &91.3$\pm$0.2 &68.0$\pm$0.4 &79.7$\pm$0.3 &95.4$\pm$0.1 &81.0$\pm$0.3 &88.2$\pm$0.2 &91.1$\pm$0.1 &67.6$\pm$0.4 &79.3$\pm$0.3 &95.3$\pm$0.1 &80.7$\pm$0.3 &88.0$\pm$0.2	\\
		
		MRResNet &full  &\textbf{94.0}$\pm$\textbf{0.1}	&\textbf{76.3}$\pm$\textbf{0.5}	&\textbf{85.2}$\pm$\textbf{0.3}	&\textbf{96.9}$\pm$\textbf{0.1}	&\textbf{86.6}$\pm$\textbf{0.3}	&\textbf{91.7}$\pm$\textbf{0.2} &\textbf{93.3}$\pm$\textbf{0.3}	&\textbf{74.8}$\pm$\textbf{0.4}	&\textbf{84.0}$\pm$\textbf{0.4}	&\textbf{96.5}$\pm$\textbf{0.2}	&\textbf{85.6}$\pm$\textbf{0.3}	&\textbf{91.1}$\pm$\textbf{0.2}
		\\
		
		MRResNet(Ours) &weak  &91.6$\pm$0.1 &69.0$\pm$0.3 &80.3$\pm$0.2 &95.6$\pm$0.1 &81.6$\pm$0.2 &88.6$\pm$0.1 &91.5$\pm$0.2 &68.3$\pm$0.5 &79.9$\pm$0.3 &95.5$\pm$0.1 &81.1$\pm$0.3 &88.3$\pm$0.2	\\
		\hline
		\hline
	\end{tabular}%
	\label{table:1}%
\end{table*}%

\subsection{Parameter Settings}
In our experiment, we use multiple mainstream models as feature extraction networks and embed the neighbor sampler and feature aggregation module, including the FCN \cite{2015Fully}, UNet \cite{2015U}, ResUNet \cite{8900625}, NestedUNet \cite{2018UNet}, D-LinkNet \cite{2018D}, DeepLab V3+ \cite{2018Encoder}, MFDeepLab V3+ \cite{8883176}, SANet \cite{9269403}, SNS-CNN \cite{8898367}, DeepUNet \cite{8573826}, and MRResNet \cite{rs13163122}. The specific settings are as follows. In the FCN model, we choose to use FCN-8s. Specifically, the last three layers of feature maps are gradually up-sampled and merged, and finally the output was obtained. The UNet model uses a four-layers encoder-decoder structure. The encoder consists of two consecutive \begin{math}3\times3 \end{math} Conv-BN-ReLU layers and a maxpooling layer. The decoder consists of two \begin{math}3\times3 \end{math} Conv-BN-ReLU layers and an upsampling layer. We connect two consecutive \begin{math}3\times3\ \end{math} Conv-BN-ReLU modules into a residual structure for ResUNet; other structures are the same as UNet. The depth of NestedUNet is set to four layers, and dense connections are added to each layer. DLinkNet uses pre-trained ResNet as the backbone. We set the size of the kernel for different dilated convolutions in Dblock to \begin{math}3\times3,\ \end{math}and the dilated rate is 6, 12, 18, and 24. For the ASPP in Deeplab V3+, we set up a point convolution, a global average pooling layer, and three dilated convolutions, with the dilated rate set to 6, 12, and 18. In MFDeepLab V3+, the weighting coefficients of three different scale features are set to 0.4, 0.3 and 0.3. In SANet, we use an adaptive multi-scale feature learning module (AML) to replace the convolution module of UNet. The SNA-CNN model uses a four-layer encoder-decoder structure. DeepUNet uses a six-layer encoder-decoder structure; for fair comparison, superpixel-based segmentation and CRF are not used. In MRResNet, the dilated rates of the Multiscale Dilated Convolution (MSDC) Module are set to 1, 2, 4, 8 and 16. The Multikernel Max Pooling (MKMP) Module contains contextual information for four different sizes of receptive fields:  \begin{math}2\times2\ \end{math}, \begin{math}3\times3\ \end{math}, \begin{math}5\times5, \end{math} and \begin{math}6\times6\ \end{math}. When the value of \begin{math} K \end{math} is 2, an additional decoder is added to make the output size consistent with the label. In other cases, the output is upsampled to the same size as the label.

In the proposed NFANet, only random horizontal flips, random up and down flips, and random rotations of 90 degrees are used for data augmentation. We set the batch size to 4. In recursive training, cross entropy plus dice loss is used as the loss function. We set the learning rate to 0.0001 and use the early stop strategy to prevent over fitting. The specific early stopping strategy is to halve the learning rate if the training loss does not decrease every 3 epochs, and terminate the training if the training loss does not decrease for 6 epochs; and the total epoch upper limit is set to 100 times. The optimizer adopts the Adam optimizer with weight decay, and the weight decay is set to 0.001.

\subsection{Comparison with Fully-Supervised Approaches}

\begin{figure*}
	\centering
	\includegraphics[width=1\linewidth]{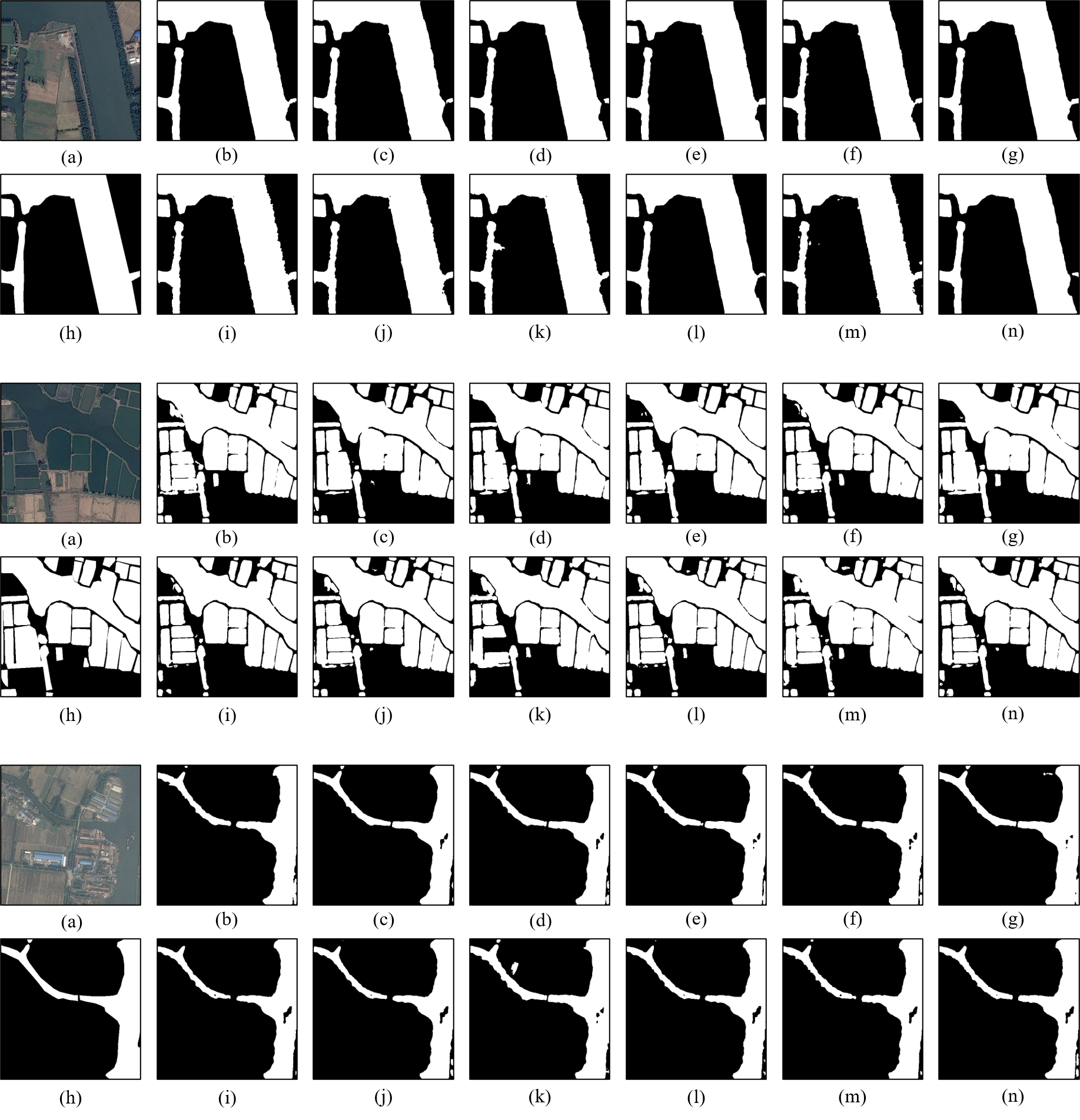}
	\caption{Prediction results of the fully-supervised methods and our methods. The visualization results of some new models are shown below. (a) and (h) represent the original image and ground truth. (b)-(g) represent the prediction results of NestedUNet, MFDeepLab V3+, SANet, SNS-CNN, DeepUNet and MRResNet, respectively. (i)-(n) represent the prediction results of our method based on NestedUNet, MFDeepLab V3+, SANet, SNS-CNN, DeepUNet and MRResNet, respectively.}
	\label{fig:7}
\end{figure*}

We compared our method with several other fully supervised semantic segmentation approaches. 
In general, datasets are randomly divided into three sets, where 60\% of the images are used for training, 20\% of the images are used for validation, and 20\% of the images are used for testing. The following 11 methods are considered for comparison:

\begin{enumerate}
{\item
Fine-tuned FCN. The feature extractor is a pre-trained VGG-Net, and we set the output dimension of the last layer to the dimension of the category.}

{\item
UNet. We use a four-layer encoder-decoder structure, and the other settings are basically the same as the weakly supervised parameter settings. UNet usually performs well in two-category semantic segmentation.}

{\item
Res-UNet. Based on an improvement of the UNet structure, we composed the convolutional layer in the encoder or decoder into residual connections.}

{\item
NestedUNet. Based on an improvement of the UNet structure, it contains more long and short connections, and it has been proved that the feature extraction ability is more powerful.}

{\item
D-LinkNet. This is the model that won first place in the 2018 CVPR road extraction competition. It was good at two-category semantic segmentation; the Dblock in the middle part continuously uses multiple dilated convolutions for dense connection.}

{\item
DeepLab V3+. The ASPP module extracts deep semantic information from different receptive fields, and uses global average pooling to obtain global features and more semantic details.}

{\item
MFDeepLab V3+. The decoder uses multi-scale features weighted fusion to further refine the details of the water area, and the weight coefficients of different features need to be set as hyperparameters.}

{\item
SANet. The model combines multi-feature extraction and adaptive feature fusion to form an AML.}

{\item
SNS-CNN. This model embeds the dilated convolution into the up-sampling operation to keep the decoding details.}

{\item
DeepUNet. The model uses a deep UNet network for water extraction, and uses superpixel-based segmentation and CRF to further refine the results.}

{\item
MRResNet. The model uses MSDC Module and MKMP Module, which helps preserve the boundaries of the water-bodies.}

\end{enumerate}

\begin{table*}
	\setlength{\abovecaptionskip}{0.pt}
	\setlength{\belowcaptionskip}{-0.em}
	\renewcommand\tabcolsep{3.5pt} 
	\centering
	\footnotesize
	\caption{Comparison of Water Extraction Results (\%) and Weak Supervision}
	\begin{tabular}{cccccccccccccc}
		\hline
		\hline 
		\multirow{2}{*}{Method}  & \multicolumn{6}{c}{Validation} & \multicolumn{6}{c}{Test}  \\
		\cmidrule(r){2-7} \cmidrule(r){8-13}
		&bgIoU &fgIoU  &mIoU &bgDice &fgDice &mDice &bgIoU &fgIoU  &mIoU &bgDice &fgDice &mDice \\
		\hline
		LocMap  &87.6$\pm$0.4 &54.5$\pm$1.0 &71.0$\pm$0.7 &93.4$\pm$0.2 &70.6$\pm$0.8 &82.0$\pm$0.5 &87.1$\pm$0.4 &53.1$\pm$0.8 &70.1$\pm$0.6 &93.1$\pm$0.2 &69.4$\pm$0.7 &81.2$\pm$0.5	\\
		SPMF-Net  &85.4$\pm$0.4 &48.5$\pm$0.6 &66.9$\pm$0.5 &92.1$\pm$0.2 &65.3$\pm$0.6 &78.7$\pm$0.4 &85.0$\pm$0.3 &47.7$\pm$0.5 &66.4$\pm$0.4 &91.9$\pm$0.2 &64.6$\pm$0.5 &78.2$\pm$0.3	\\
		U-CAM  &86.8$\pm$0.4 &52.1$\pm$1.0 &69.5$\pm$0.7 &92.9$\pm$0.2 &68.5$\pm$0.9 &80.7$\pm$0.6 &86.3$\pm$0.4 &51.0$\pm$0.8 &68.7$\pm$0.6 &92.7$\pm$0.2 &67.5$\pm$0.7 &80.1$\pm$0.5	\\
		Baseline   &41.1$\pm$5.9 &14.2$\pm$5.8 &27.7$\pm$5.5 &58.1$\pm$5.8 &24.5$\pm$8.8 &41.3$\pm$6.9 &41.4$\pm$5.9 &14.1$\pm$5.8 &27.7$\pm$5.6 &58.3$\pm$5.8 &24.4$\pm$8.8 &41.4$\pm$7.0 \\
		\hline
		Ours  &\textbf{91.4}$\pm$\textbf{0.4} &\textbf{67.8}$\pm$\textbf{1.3} &\textbf{79.6}$\pm$\textbf{0.8} &\textbf{95.5}$\pm$\textbf{0.2} &\textbf{80.8}$\pm$\textbf{0.9} &\textbf{88.2}$\pm$\textbf{0.5} &\textbf{91.1}$\pm$\textbf{0.3} &\textbf{67.0}$\pm$\textbf{0.7} &\textbf{79.1}$\pm$\textbf{0.5} &\textbf{95.3}$\pm$\textbf{0.2} &\textbf{80.3}$\pm$\textbf{0.5} &\textbf{87.8}$\pm$\textbf{0.3}	\\
		\hline
		\hline
	\end{tabular}%
	\label{table:2}%
\end{table*}%

\begin{figure*}
	\centering
	\includegraphics[width=1\linewidth]{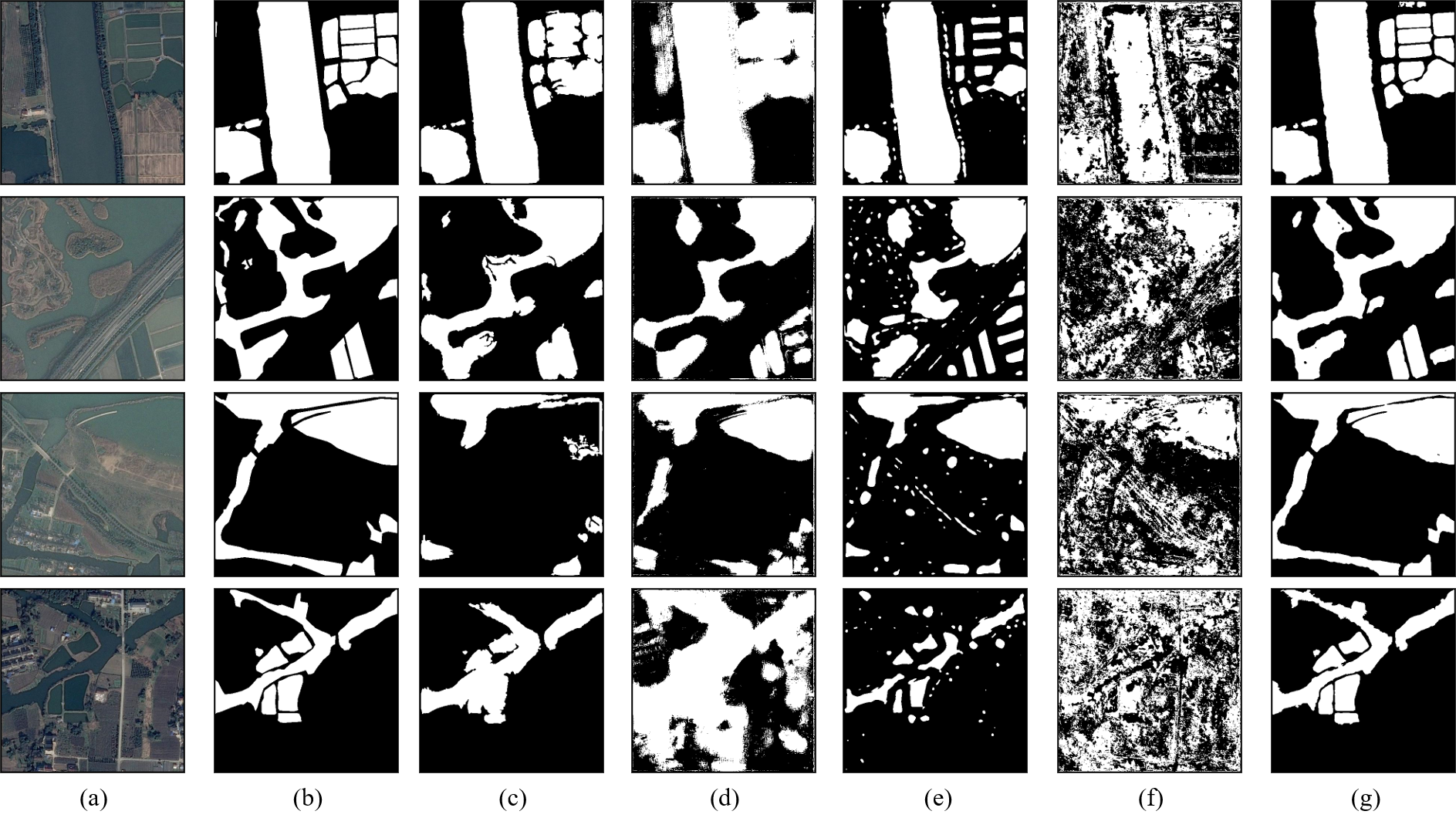}
	\caption{Prediction results of the other weakly supervised methods and our method. (a) and (b) represent the original image and ground truth. (c) represents Loc-Maps method.   (d) represents SPMF-Net. (e) represents U-CAM. (f) represents baseline. (g) represents our method.}
	\label{fig:8}
\end{figure*}

In Table \ref{table:1}, we report the water extraction performance of the proposed approach and compare it with the fully-supervised approaches. The best scores are shown in bold in the table. Fig. \ref{fig:7} also illustrates the visual performance of all approaches. Experiments demonstrate that our method achieves the best score together with the NestedUNet-based model. The visual performance showes that the prediction results obtained by our method are very close to the ground truth. The bgIoU of our method reaches 91.6\%, the fgIoU reaches 68.8\%, the mIoU reaches 80.2\%, the bgDice reaches 95.6\%, the fgDice reaches 81.5\%, and the mDice reaches 88.6\%. Compared with the best fully-supervised model MRResNet, the mIoU of our method is only reduced by 3.8 percentage points, and mDice is only reduced by 2.5 percentage points. However, the labeling cost of our method is much less than that of the fully-supervised method. From Fig. \ref{fig:7}, we can observe the following results. The edges of the weakly-supervised prediction results are not very smooth compared to the fully-supervised prediction results, and there are still some defects in some local small water-bodies. We observe that these incomplete small water-bodies are more difficult to identify, and are usually  mistaken for grass or bare soil. This is likely to be the reason for the gap between our results and the results of the fully-supervised method. For the impact of the amount of training data, we conduct a more detailed ablation experiment ( see Table IV and Fig. 12).

\subsection{Comparison with Weakly Supervised Approaches}

We compared our method with several other weakly supervised remote sensing approaches. Data sets are randomly divided into three sets, where 60\% of the images are used for training, 20\% of the images are used for validation, and 20\% of the images are used for testing. The other weak supervision methods used for comparison are as follows.

\begin{enumerate}

{\item
In \cite{2018WSF}, the author uses image-level labels supervision to generate localization maps, combined with traditional algorithms based on graph algorithms to obtain pseudo-labels, and then uses pseudo-label supervision to train the network. For fair comparison, we use point labels for supervision, and then CMax pooing to generate localization maps.}

{\item
In \cite{2020SPMF}, the model uses image-level labels for supervision. We remove the global average pooling layer and the fully connected layer, and used point labels for supervision.}

{\item
In \cite{rs12020207}, the author combines UNet and CAM to form a U-CAM. Specifically, U-CAM is obtained by the weightes sum of the filter dimensions output by the last convolutional layer, where the weight comes from the fully connected layer. Then threshold processing is performed on U-CAM to obtain hard segmentation prediction.}

{\item
We train a UNet that only uses point labels for supervision. The loss function only calculates positive labels, and negative labels do not participate in the loss calculation. We use it as the baseline of point supervision.}

\end{enumerate}

The experimental results of the comparison of our method with other weakly supervised remote sensing approaches are shown in Table \ref{table:2}. Fig. \ref{fig:8} shows the prediction results of other weakly supervised methods in comparison to the proposed method. To be fair, all methods are based on UNet. The bgIoU of our method reaches 91.1\%, the fgIoU reaches 67.0\%, the mIoU reaches 79.1\%, the bgDice reaches 95.3\%, the fgDice reaches 80.3\%, and the mDice reaches 87.8\%. Compared with the best weakly supervised method, LocMap, the mIoU of our method improves 9.0 percentage points, and mDice improves 6.6 percentage points. Although other weak supervision methods can predict the local area of the water-body, there are errors in the detection of the water-body boundary, while our method is relatively more accurate. The compared weak supervision methods cannot detect small objects appropriately, while this issue is largely solved to by the proposed method. If only point labels are used as positive labels to train the network, it is difficult for the network to predict water-bodies and non-water-bodies.

\subsection{Ablation Experiments}

The method proposed in this paper mainly includes two modules and one strategy, i.e., the neighbor sampler module, the feature aggregation module, and the recursive training strategy. In order to prove the effectiveness of each module or strategy, ablation experiments were performed on the data-set.

\subsubsection{Effectiveness of neighbor sampling}

\begin{figure}
	\centering
	\includegraphics[width=1\linewidth]{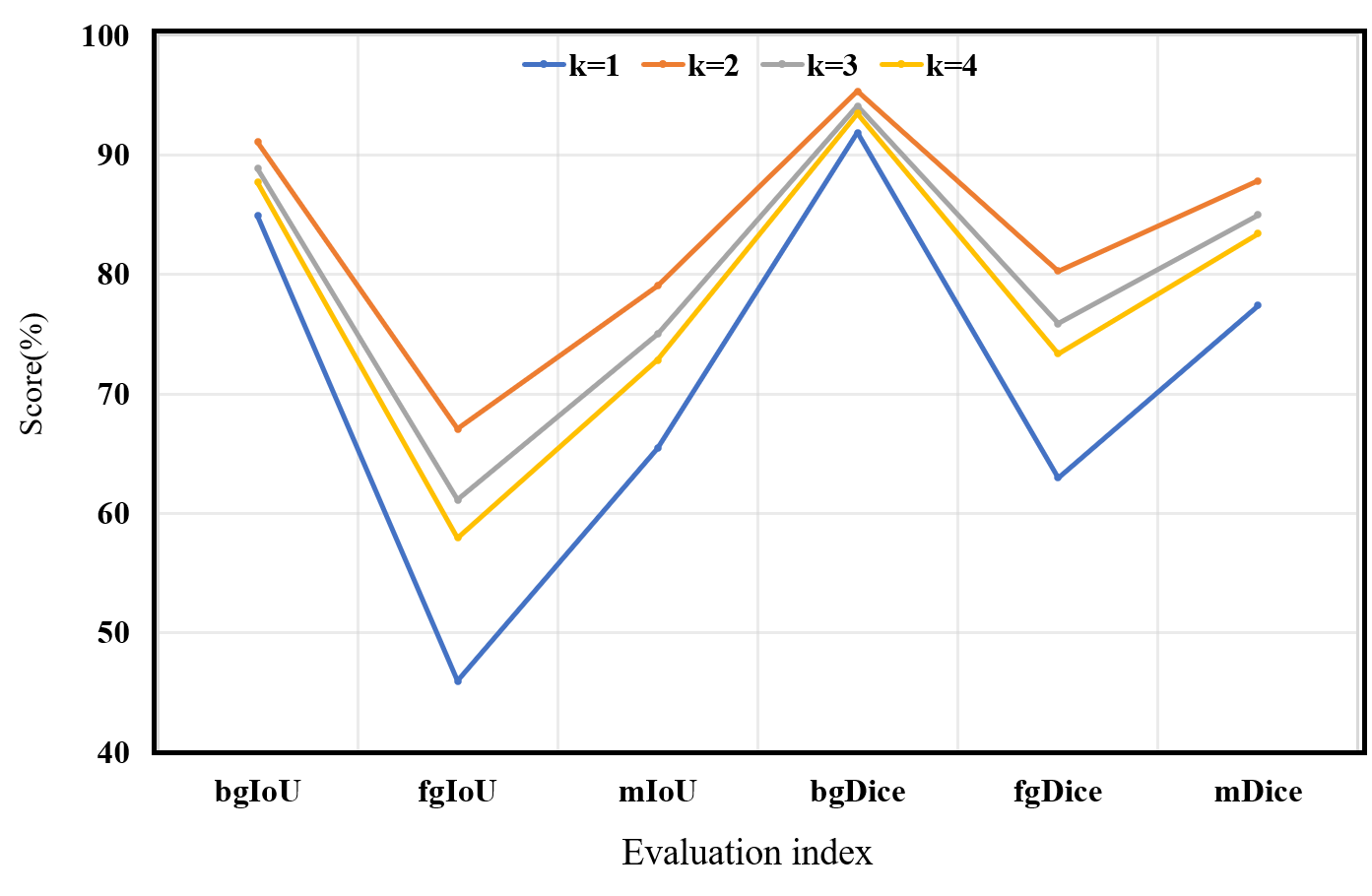}
	\caption{Effectiveness of neighbor sampling.}
	\label{fig:9}
\end{figure}

\begin{table*}
	\setlength{\abovecaptionskip}{0.pt}
	\setlength{\belowcaptionskip}{-0.em}
	\centering
	\normalsize
	\caption{Effectiveness of Feature Aggregation(\%)}
	\begin{threeparttable}
	\begin{tabular}{cccccccccccccc}
		\hline
		\hline 
		NF\tnote{1} &RT\tnote{2} &bgIoU &fgIoU  &mIoU &bgDice &fgDice &mDice  \\
		\hline
		\begin{math} f_1 \end{math} &    &84.08 &39.78 &61.93 &91.35 &56.92 &74.13 \\
		\begin{math} f_1 \end{math} &\checkmark  &87.45 &42.01 &64.73 &93.30 &59.17 &76.23 \\
		\begin{math} f_2 \end{math} &    &84.31 &49.33 &66.82 &91.49 &66.07 &78.78 \\
		\begin{math} f_2 \end{math} &\checkmark  &88.98 &51.57 &70.27 &94.17 &68.04 &81.11\\
		\begin{math} f_3 \end{math} &            &83.55 &48.37 &65.96 &91.04 &65.21 &78.12 \\
		\begin{math} f_3 \end{math} &\checkmark  &87.52 &51.48 &69.50 &93.35 &67.97 &80.66 \\
		\begin{math} f_4 \end{math} &            &82.21 &45.75 &63.98 &90.24 &62.78 &76.51 \\
		\begin{math} f_4 \end{math} &\checkmark  &87.27 &48.09 &67.68 &93.20 &64.95 &79.07 \\
		\begin{math} \left[f_1:f_2:f_3:f_4\right]\ \end{math} &  &89.63 &63.73 &76.68 &94.53 &77.85 &86.19 \\
		\begin{math} \left[f_1:f_2:f_3:f_4\right]\ \end{math} &\checkmark &\textbf{91.40} &\textbf{67.73} &\textbf{79.56} &\textbf{95.51} &\textbf{80.76} &\textbf{88.13} \\
		\hline
		\hline
	\end{tabular}%
         \begin{tablenotes}    
		\footnotesize               
		\item[1] “NF” Means Neighbor Features. 
		\item[2] “RT” Means Using Recursive Training.
\end{tablenotes}
	\end{threeparttable}
	\label{table:3}%
\end{table*}%

In the ablation experiment, the other settings are unchanged, and only the value of \begin{math} K\end{math} is changed. We set the neighbor sampling parameter \begin{math} K\end{math} of our proposed network from 1 to 4 and only use cross entropy and dice loss to train the model. For different \begin{math} K\end{math} values of NFANet, in order to avoid interference from other modules, we only select UNet as the feature extraction network for comparative experiments. In particular, when the value of \begin{math} K\end{math} is set to 1, the neighbor image group degenerates into the input image. As shown in Fig. \ref{fig:9}, with the gradual increase of \begin{math} K\end{math}, mIoU first increases and then decreases. As the neighbor sampling parameter \begin{math} K\end{math} increases, the number of adjacent pixels that need to be considered increase geometrically, resulting in information redundancy, and the size of each reconstructed neighbor image is gradually reduced. Therefore, we set \begin{math} K\end{math} equal to 2, because the neighbor features require less computation and achieves better performance.

\subsubsection{Effectiveness of feature aggregation} 

\begin{table*}
	\setlength{\abovecaptionskip}{0.pt}
	\setlength{\belowcaptionskip}{-0.em}
	\centering
	\small
	\caption{Segmentation Results(\%) Using Different Proportions of The Training Set}
	\begin{tabular}{ccccccccccccc}
		\hline
		\hline 
		\multirow{2}{*}{Method}  & \multicolumn{2}{c}{Train=100} & \multicolumn{2}{c}{Train=200} & \multicolumn{2}{c}{Train=300}
		& \multicolumn{2}{c}{Train=400} & \multicolumn{2}{c}{Train=500} & \multicolumn{2}{c}{Train=600} \\
		\cmidrule(r){2-3} \cmidrule(r){4-5} \cmidrule(r){6-7} \cmidrule(r){8-9} \cmidrule(r){10-11} \cmidrule(r){12-13}
		&mIoU &mDice &mIoU &mDice &mIoU &mDice &mIoU &mDice &mIoU &mDice &mIoU &mDice  \\
		\hline
		UNet   &80.5	&88.7	&81.0 	&89.1 	&82.2  &89.9 &82.6	&90.1	&82.7 	&90.2 	&\textbf{83.1}  &\textbf{90.5}	\\
		Ours   &66.9	&78.9	&69.1 	&80.8 	&72.8  &83.4 &74.8	&84.7	&76.6 	&86.1 	&\textbf{79.5}  &\textbf{88.1}	\\
		\hline
		\hline
	\end{tabular}%
	\label{table:4}%
\end{table*}%

\begin{math} K\end{math} is set to 2, and the pseudo-label is obtained separately for each image in the neighbor image group (Eqs. (1) and (2)). As shown in Table \ref{table:3}, when the features of the  \begin{math} i-th \end{math} neighbor image are \begin{math} f_i \end{math}, \begin{math} \left[f_1:f_2:f_3:f_4\right]\ \end{math}, then feature aggregation is used. “RT” means that a recursive training strategy is used. Compared with the result of the best single neighbor image, the neighbor feature aggregation module improves mIoU and mDice by 9.29 percentage points and 7.02 percentage points, respectively. Compared with the worst single neighbor image result, the neighbor feature aggregation module improves mIoU and mDice by 14.83 percentage points and 11.90 percentage points, respectively. Due to the different contributions of neighbor pixels to prediction results, the scores of the pseudo-labels generated by a single neighbor image are different. The visualization results after using CMax pooling are shown in Fig. \ref{fig:4}. The feature aggregation module uses the features extracted from neighbor pixels to make joint judgments, and the information of different neighbor pixels complement each other, so it generates more high-quality pseudo-labels and ultimately improves the score of the water-bodies prediction result.

\subsubsection{Effectiveness of recursive training} 

\begin{figure}
	\centering
	\includegraphics[width=1\linewidth]{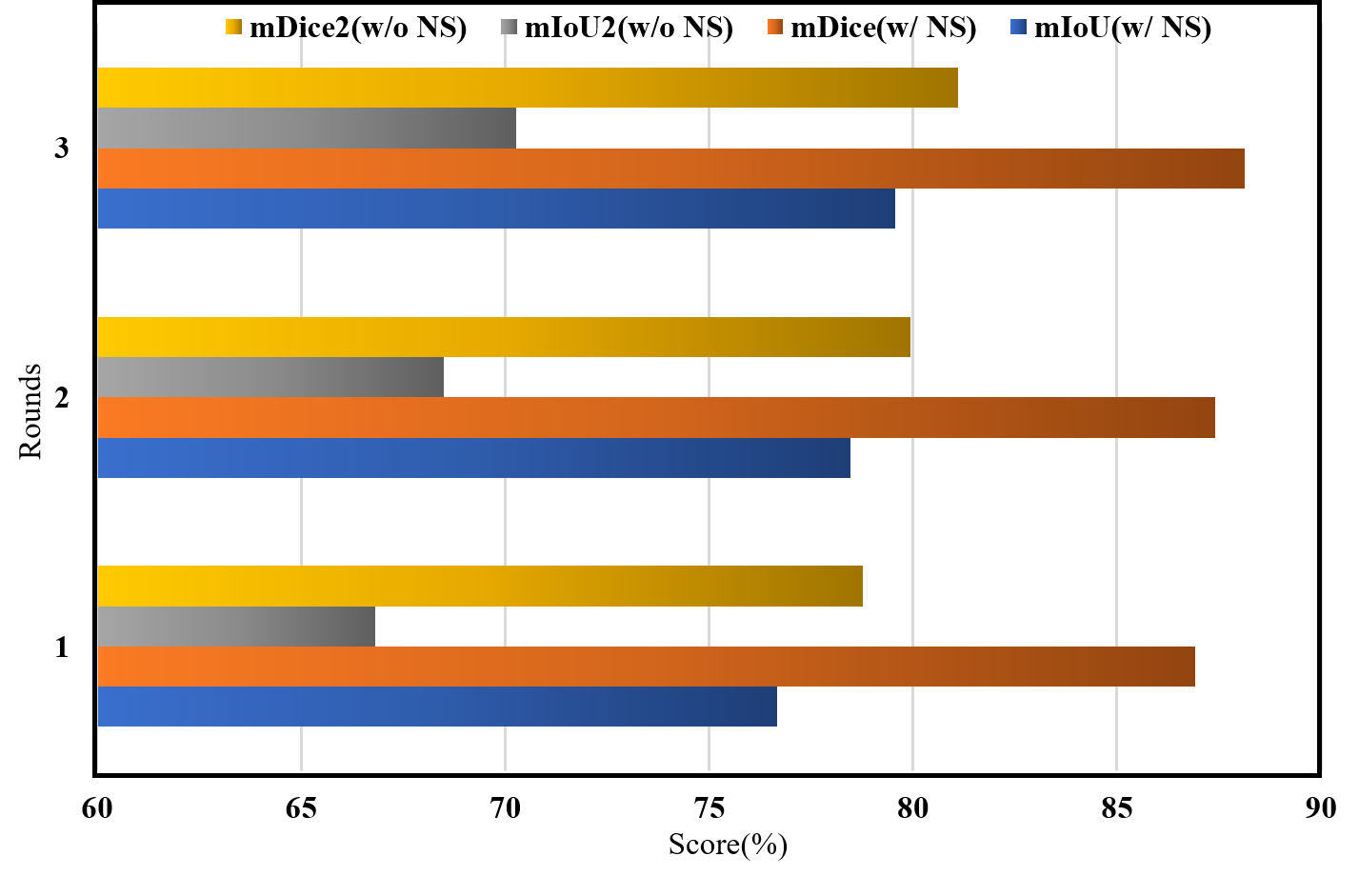}
	\caption{Effectiveness of recursive training.}
	\label{fig:10}
\end{figure}

\begin{figure}
	\centering
	\includegraphics[width=1\linewidth]{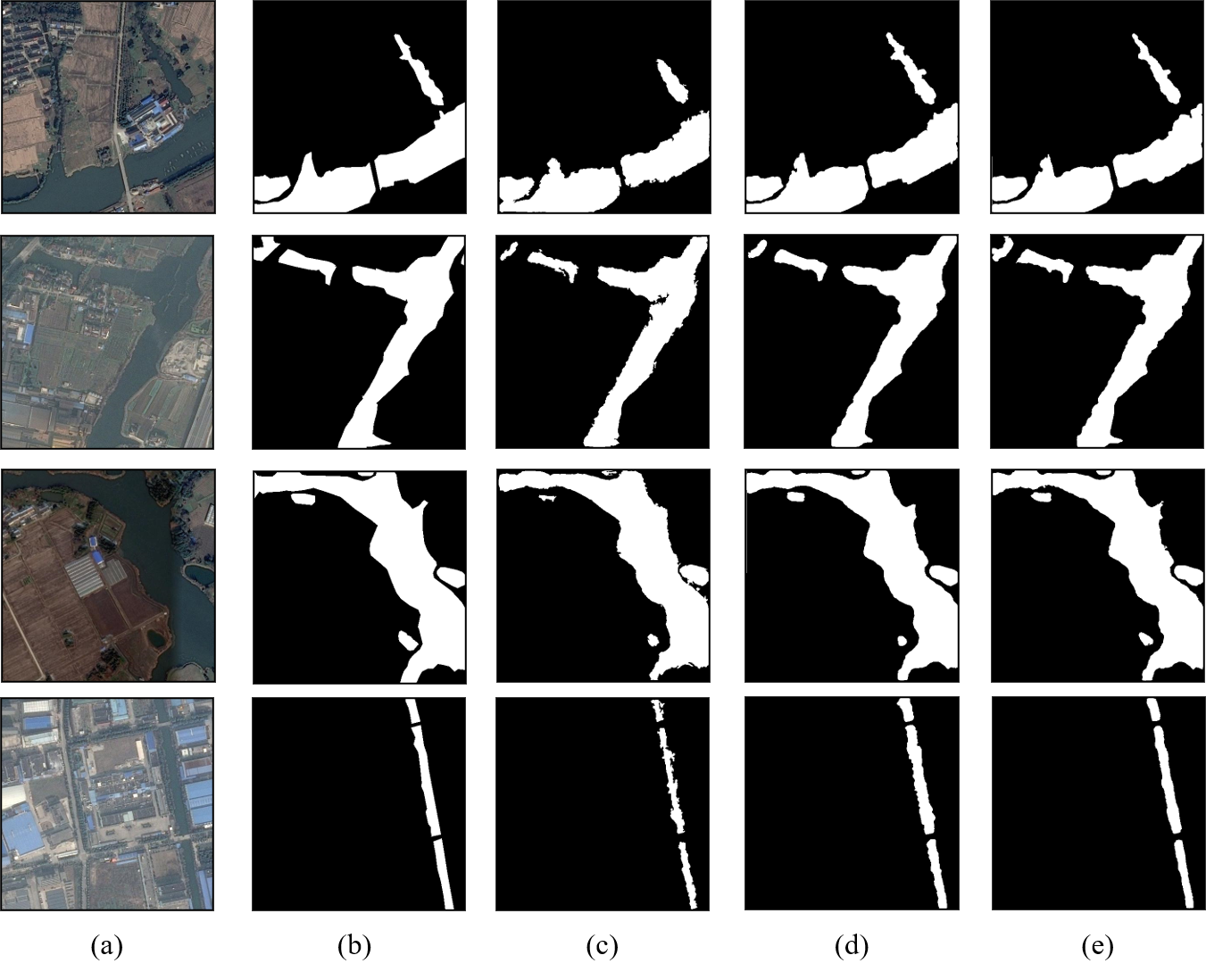}
	\caption{Predicted results of recursive training. (a) and (b) represent the original image and ground truth. (c)-(e) represents the results of the first, second and third rounds of recursive training respectively.}
	\label{fig:11}
\end{figure}

We set recursion times to 3 and compare the evaluation indicators of each recursive training. As shown in Fig. \ref{fig:10}, “w/o NS” means neighbor sampling  was not used. As can be seen from the figure, whether using neighbor sampling or not, mIoU and mDice are improved. Compared with the preliminary results, mIoU and mDice increased by 2.88 percentage points and 1.22 percentage points, respectively, after using recursive training. Fig. \ref{fig:11} shows the visualization results of three rounds of recursion. It can be seen from the figure that there are many burrs on the water boundary in the first round, and there are still some errors in the local small water-bodies. After recursive training, the ability of the network to capture the water contour is improved, and the water boundary is gradually smoothed. The boundaries between water-bodies are also clearer, even the small boundaries that are mislabeled in ground truth are correctly distinguished.

\subsubsection{Impact of data split} 
\begin{figure}
	\centering
	\includegraphics[width=1\linewidth]{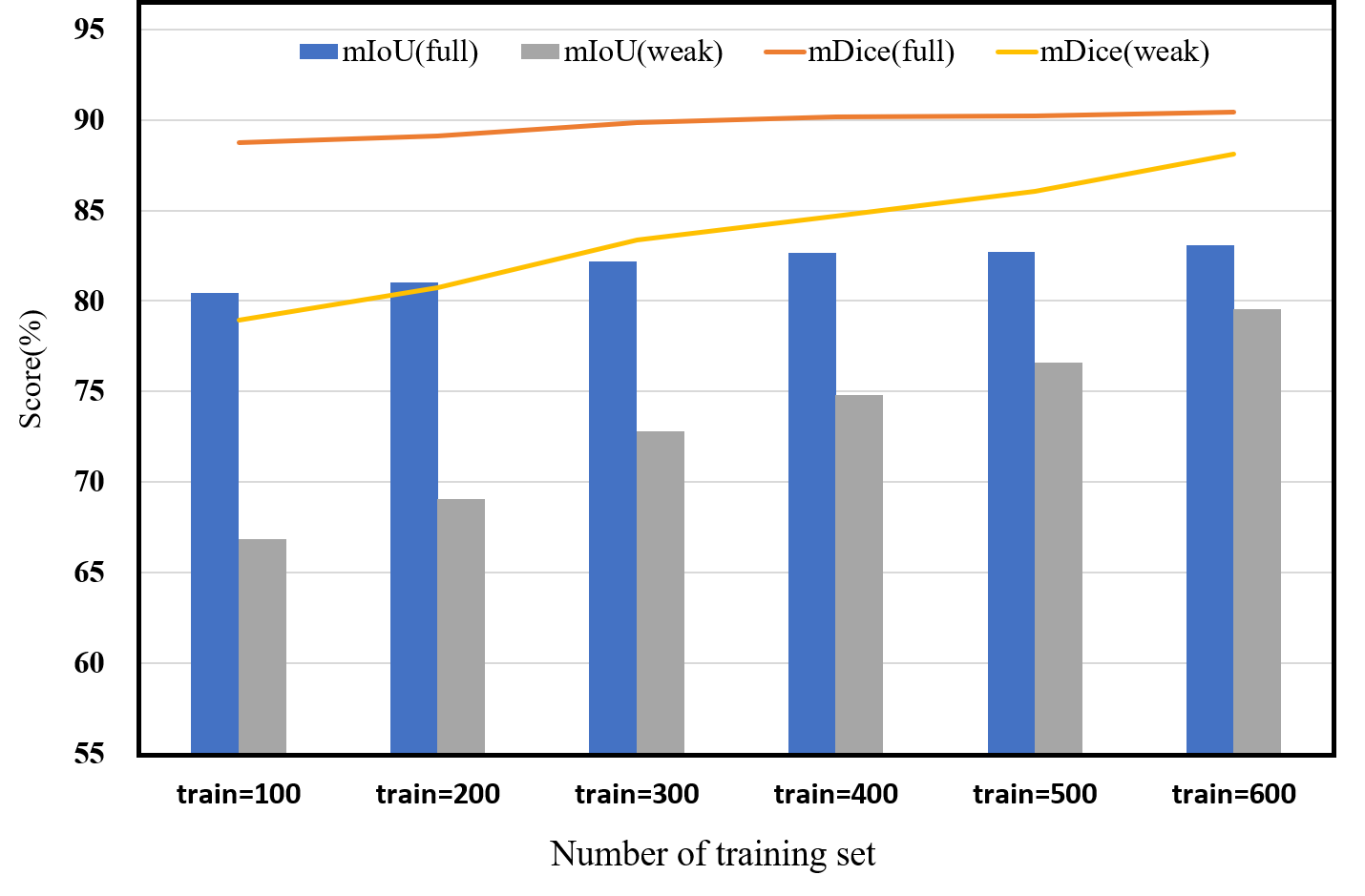}
	\caption{Impact of data split.}
	\label{fig:12}
\end{figure}

This section explores the impact of the data split. We compare the fully-supervised method with our method (Table \ref{table:4}). When the amount of data in the training set is increased from 100 to 600, the fully-supervised mIoU score increases by 2.6 percentage points, and the mIoU score of our method increases by 12.6 percentage points. The following conclusions can be drawn from the Fig. \ref{fig:12}. As the amount of data gradually increases, both the fully-supervised method and our method are improved. When the amount of data is small, the point-level label seriously lacks the supervision information that it can provide, the neighbor features are not enough to distinguish water-bodies from non-water-bodies, and the generated pseudo-labels have much wrong information. The supervised information provided by pixel-level labels is more comprehensive, which enables the neural network to learn some key features. When the amount of data increases, neighbor features play an increasingly important role, which enables the neural network to better distinguish water-bodies from remote sensing images, and the quality of pseudo-labels is significantly improved.

\subsection{Time Consumption and Model Parameters}

\begin{table*}
	\setlength{\abovecaptionskip}{0.pt}
	\setlength{\belowcaptionskip}{-0.em}
	\centering
	\scriptsize
	\caption{GPU Inference Time and Model Parameters}
	\begin{tabular}{cccccccccccc}
		\hline
		\hline 
		Method &FCN &UNet &Res-UNet &NestedUNet &D-LinkNet &DeepLab V3+ &MFDeepLab V3+ &SANet &SNS-CNN &DeepUNet &MRResNet \\
		\hline
		Times(MS) &19.0 &16.2 &18.9 &35.7 &\textbf{12.9} &26.5 &75.7 &34.2 &25.0 &15.5 &28.4 \\
		Param.(MB) &74.6 &31.5 &33.2 &36.7 &124.5 &161.7 &485.1 &\textbf{13} &62.9 &39.2 &278.7 \\
		\hline
		\hline
	\end{tabular}%
	\label{table:5}%
\end{table*}%

The hardware configurations for the experiments in this paper consisted of Intel Core i7-9700k 3.60 GHz CPU, GeForce RTX 2080Ti GPU, and 16GB RAM. The results of the GPU inference time, shown in Table \ref{table:5}, are the average GPU inference time of the data-set. After using recursive training to improve the quality of pseudo-labels, we input the pseudo-labels and original images into the models that are consistent with the fully-supervised methods for training and inference. Therefore, the GPU inference time of the proposed method is the same as that of the fully-supervised methods. It can be observed from the table that the inference time of D-LinkNet is the shortest. This is because D-LinkNet compresses the feature channel in the decoder to reduce the computational cost. MFDeepLab V3+ takes the longest time, due to its use of multi-scale feature weighted fusion, which requires multiple DeepLab V3+ models for prediction.

The above methods are divided into two categories: UNet-based models and ResNet-based models, and so the model parameters are different. In the UNet-based models, the parameter of SANet is the smallest at 13 MB, and the parameter of SCS-CNN is the greatest at 62.9 MB. In the ResNet-based model, the parameter of FCN is the smallest at 74.6 MB, and the parameter of MFDeepLab V3+ is the greatest at 485.1 MB.

\subsection{Hard case}

\begin{figure}
	\centering
	\includegraphics[width=1\linewidth]{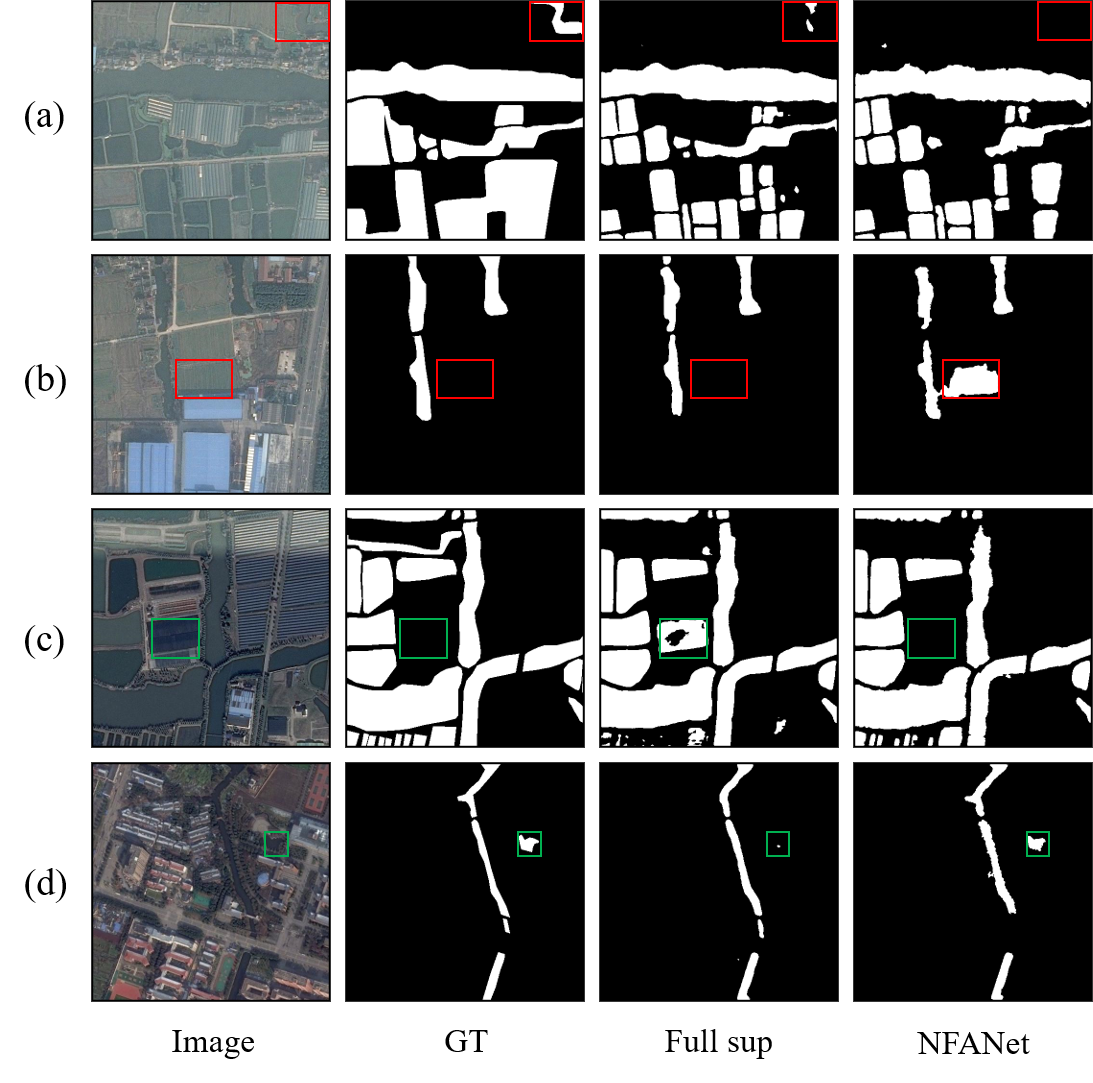}
	\caption{Impact of data split.}
	\label{fig:13}
\end{figure}

Although our method yields good results, there are still some errors. As shown in Figure 13, the entire water body area in the red box in Fig. \ref{fig:13}a has missed predictions. We think that the border between the cultivated land on its left and the water body is fuzzy and difficult to distinguish. The cultivated land in the red box in Fig. \ref{fig:13}b is wrongly predicted to become a water body. It is worth noting that in some images that are difficult to distinguish water bodies, the proposed method is better than full supervision. As shown in the green box of Fig. \ref{fig:13}c, the fully supervised model incorrectly predicted the greenhouse as a water body. A small piece of water in Fig. \ref{fig:13}d is not predicted, and our method can find it well.

\section{Conclusion}
In this paper, we proposed a network entitled NFANet. Unlike traditional convolutional neural networks that only use global or local features for discrimination, NFANet uses neighbor features, which allows more representative features to be learned. We propose a method called neighbor sampling to obtain neighboring images. By using the feature aggregation module, we fuse these neighbor features to obtain pseudo-labels, and improve the label quality by recursive training. We tested it on water data sets and compared it with advanced fully supervised and weakly supervised methods. By using only point labels, the proposed method obtains comparable results with that of full supervision. 

As a possible future work, we will conduct research on weakly supervised or semi-supervised methods of self-correction. Moreover, we will also consider the characteristics of other different ground objects and design a dedicated module for feature extraction to further improve the ground object extraction performance of CNN.


\ifCLASSOPTIONcaptionsoff
  \newpage
\fi

\bibliography{NFANet}

\bibliographystyle{IEEEtran}

\begin{IEEEbiography}[{\includegraphics[width=1in,height=1.25in,clip,keepaspectratio]{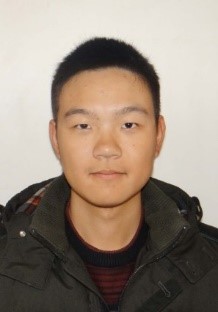}}]{Ming Lu} was born in Hubei, China, in 1996. He received the B.S. degree from the Hubei University of Technology, Wuhan, China, in 2019. He is currently pursuing the masters degree with the College of Electrical and Information Engineering, Hunan University, Changsha. His research interests focus on weakly supervised learning and remote sensing images semantic segmentation.
\end{IEEEbiography}

\begin{IEEEbiography}[{\includegraphics[width=1in,height=1.25in,clip,keepaspectratio]{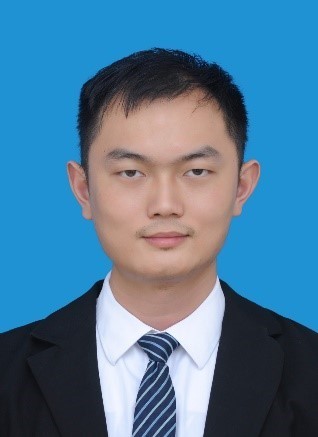}}]{Leyuan Fang} received the Ph.D. degree from the College of Electrical and Information Engineering, Hunan University, Changsha, China, in 2015. From September 2011 to September 2012, he was a Visiting Ph.D. Student with the Department of Ophthalmology, Duke University, Durham, NC, USA, supported by the China Scholarship Council. From August 2016 to September 2017, he was a Post-Doctoral Researcher with the Department of Biomedical Engineering, Duke University, Durham, NC, USA. He is a Professor with the College of Electrical and Information Engineering, Hunan University, and an Adjunct Researcher with the Peng Cheng Laboratory, Shenzhen, China. His research interests include sparse representation and multiresolution analysis in remote sensing and medical image processing. Dr. Fang was a recipient of one 2nd-Grade National Award at the Nature and Science Progress of China in 2019. He is an Associate Editor of the IEEE TRANSACTIONS ON IMAGE PROCESSING, the IEEE TRANSACTIONS ON GEOSCIENCE AND REMOTE SENSING, the IEEE TRANSACTIONS ON NEURAL NETWORKS AND LEARNING SYSTEMS, and Neurocomputing.
\end{IEEEbiography}

\begin{IEEEbiography}[{\includegraphics[width=1in,height=1.25in,clip,keepaspectratio]{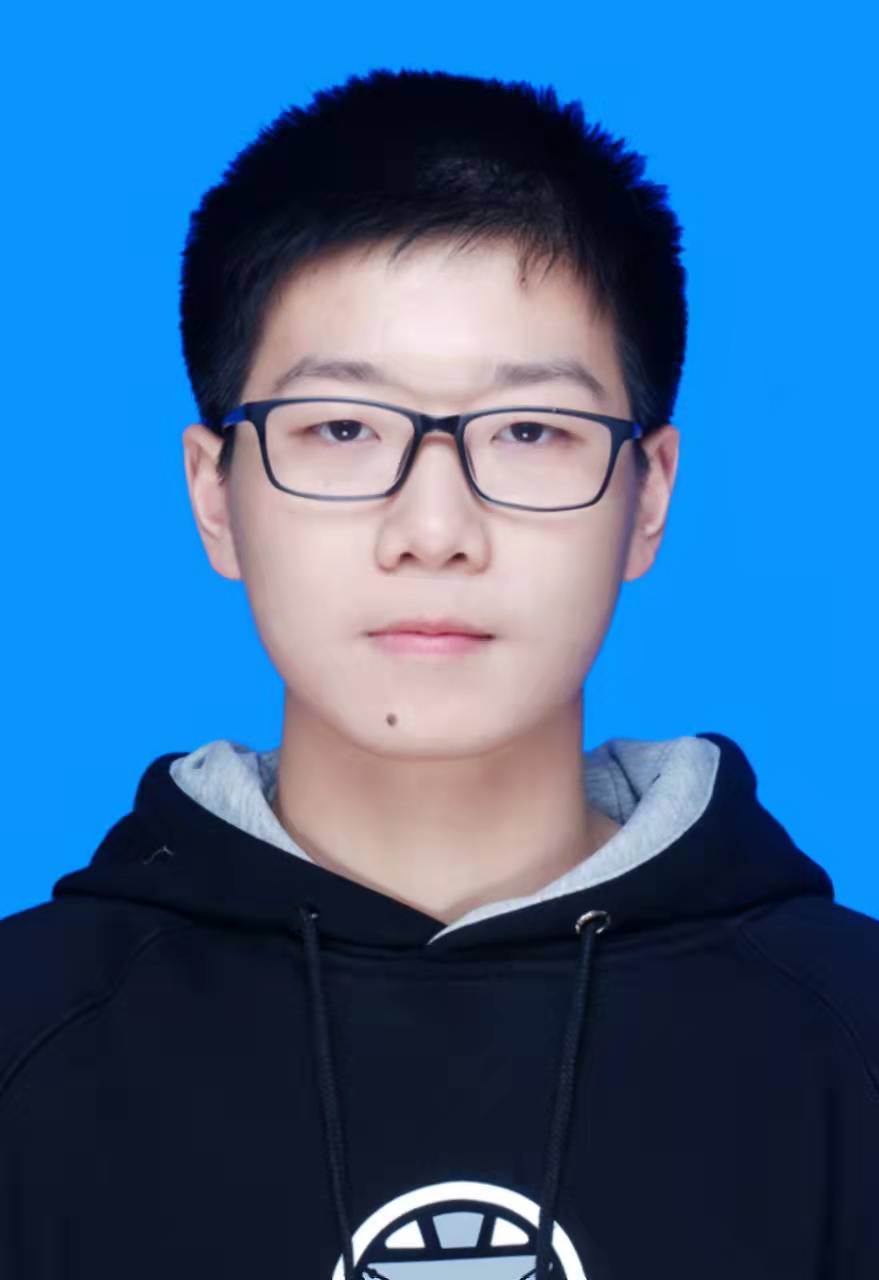}}]{Muxing Li} was born in Hunan, china, in 1999. He received a bachelor of advanced computing(Honours) degree from The Australian National University, Canberra, Australia. He is currently preparing to pursue a masters of philosophy degree with The University of Sydney, Australia. His research interests focus on computer vision.
\end{IEEEbiography}

\begin{IEEEbiography}[{\includegraphics[width=1in,height=1.25in,clip,keepaspectratio]{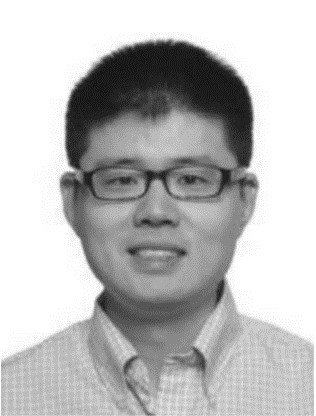}}]{Bob Zhang} (Senior Member, IEEE) received the Ph.D. degree in electrical and computer engineering from the University of Waterloo, Waterloo, ON, Canada, in 2011. He was with the Center for Pattern Recognition and Machine Intelligence and later was a Postdoctoral Researcher with the Department of Electrical and Computer Engineering, Carnegie Mellon University, Pittsburgh, PA, USA. He is currently an Associate Professor with the Department of Computer and Information Science, University of Macau, Macau, China. His current research interests include medical biometrics, pattern recognition, and image processing.
\end{IEEEbiography}

\begin{IEEEbiography}[{\includegraphics[width=1in,height=1.25in,clip,keepaspectratio]{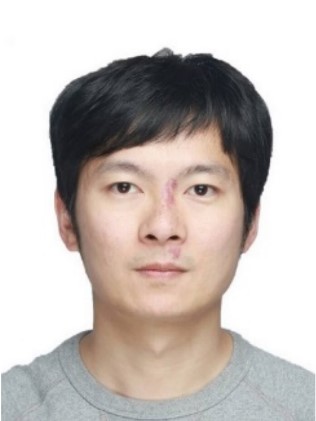}}]{Yi Zhang} received his B.S., M.S., and Ph.D. degrees from the College of Computer Science, Sichuan University, Chengdu, China, in 2005, 2008,and 2012, respectively. From 2014 to 2015, he was with the Department of Biomedical Engineering, Rensselaer Polytechnic Institute, Troy, NY, USA, as a Postdoctoral Researcher. He is currently a Full Professor with the College of Computer Science, Sichuan University, and is the director of the deep imaging group (DIG). His research interests include medical imaging, compressive sensing, and deep learning. He authored more than 70 papers in the fifield of image processing. These papers were published in several leading journals, including IEEE Transactions on Medical Imaging, IEEE Transactions on Computational Imaging, Medical Image Analysis, European Radiology, Optics Express, etc., and reported by the Institute of Physics (IOP) and during the Lindau Nobel Laureate Meeting. He received major funding from the National Key R\&D Program of China, the National Natural Science Foundation of China, and the Science and Technology Support Project of Sichuan Province, China. He is a Guest Editor of the International Journal of Biomedical Imaging, Sensing and Imaging, and an Associate Editor of IEEE Access. He is a Senior Member of IEEE.
\end{IEEEbiography}

\begin{IEEEbiography}[{\includegraphics[width=1in,height=1.25in,clip,keepaspectratio]{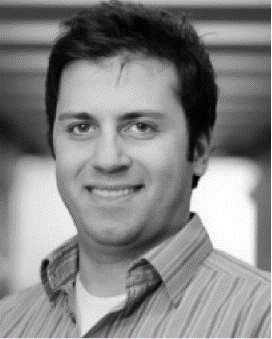}}]{Pedram Ghamisi} (S’12, M’15, SM’18) graduated with a Ph.D. in electrical and computer engineering at the University of Iceland in 2015. He works as (1) the head of the machine learning group at Helmholtz-Zentrum Dresden-Rossendorf (HZDR), Germany and (2) visiting professor and group leader of AI4RS at the Institute of Advanced Research in Artificial Intelligence (IARAI), Austria. He is a co-founder of VasoGnosis Inc. with two branches in San Jose and Milwaukee, the USA.

He was the co-chair of IEEE Image Analysis and Data Fusion Committee (IEEE IADF) between 2019 and 2021. Dr. Ghamisi was a recipient of the IEEE Mikio Takagi Prize for winning the Student Paper Competition at IEEE International Geoscience and Remote Sensing Symposium (IGARSS) in 2013, the first prize of the data fusion contest organized by the IEEE IADF in 2017, the Best Reviewer Prize of IEEE Geoscience and Remote Sensing Letters in 2017, and the IEEE Geoscience and Remote Sensing Society 2020 Highest-Impact Paper Award. His research interests include interdisciplinary research on machine (deep) learning, image and signal processing, and multisensor data fusion. He is an associate editor of IEEE JSTARS and IEEE GRSL. For detailed info, please see http://pedram-ghamisi.com/.
\end{IEEEbiography}

\end{document}